\documentclass[journal,twocolumn,10pt]{IEEEtran}
\usepackage{comment}
\usepackage{graphics}
\usepackage{graphicx}
\usepackage{subcaption}
\usepackage{amsmath} 
\usepackage{amsfonts}
\usepackage{xcolor}
\usepackage{multirow}
\usepackage{bbm, dsfont}
\usepackage{booktabs}

\ifCLASSINFOpdf
\else
\fi

\begin{document}

\title{Exploring the Boundaries of On-Device Inference: When Tiny Falls Short, Go Hierarchical}

\author{Adarsh Prasad Behera, Paulius Daubaris, Iñaki Bravo, José Gallego, Roberto Morabito, Joerg Widmer, and Jaya Prakash Champati\thanks{A.P. Behera is with KTH Royal Institute of Technology, Sweden. P. Daubaris is with the University of Helsinki, Finland. I. Bravo, J. Gallego, and J. Widmer are with the IMDEA Networks Institute, 
 Spain. R. Morabito is with EURECOM, France. J. P. Champati is with the University of Victoria, Canada.}
}


\maketitle

\begin{abstract}
%The Hierarchical Inference (HI) paradigm has recently emerged as an effective method for balancing inference accuracy, data processing, transmission throughput, and offloading cost. This approach proves particularly efficient in scenarios involving resource-constrained edge devices like micro controller units (MCUs), tasked with executing tinyML inference. Notably, it outperforms strategies such as local inference execution, inference offloading, and split inference (i.e., inference execution distributed between two endpoints). In this work, we propose 
On-device inference offers significant benefits in edge ML systems, such as improved energy efficiency, responsiveness, and privacy, compared to traditional centralized approaches. However, the resource constraints of embedded devices limit their use to simple inference tasks, creating a trade-off between efficiency and capability. In this context, the Hierarchical Inference (HI) system has emerged as a promising solution that augments the capabilities of the local ML by offloading selected samples to an edge server/cloud for remote ML inference.
%In HI, all the data samples first receive local ML inference, and only those data samples for which the inference is likely incorrect are forwarded to an edge server/cloud for remote inference. 
%Existing works use simulation to demonstrate that HI improves accuracy but do not account for the latency and energy consumption on the device. In contrast, in this paper, we systematically compare the performance of HI with on-device inference based on measurements of accuracy, latency, and energy for running embedded ML models on five devices and three image classification datasets. We demonstrate that HI systems can be designed to achieve lower latency and device energy consumption than an on-device inference system for a given accuracy requirement. 
Existing works, primarily based on simulations, demonstrate that HI improves accuracy. However, they fail to account for the latency and energy consumption in real-world deployments, nor do they consider three key heterogeneous components that characterize ML-enabled IoT systems: hardware, network connectivity, and models.
To bridge this gap, this paper systematically evaluates HI against standalone on-device inference by analyzing accuracy, latency, and energy trade-offs across five devices and three image classification datasets. Our findings show that, for a given accuracy requirement, the HI approach we designed achieved up to $73$\% lower latency and up to $77$\% lower device energy consumption than an on-device inference system. 
%The key to building an efficient HI system is the availability of on-device models with a high accuracy-to-size ratio. 
%The key to building an efficient HI system is the availability of small-size, reasonably accurate on-device models whose outputs can be effectively differentiated for samples that require remote inference. 
Despite these gains, HI introduces a fixed energy and latency overhead from on-device inference for all samples. To address this, we propose a hybrid system called Early Exit with HI (EE-HI) and demonstrate that, compared to HI, EE-HI reduces the latency up to $59.7$\% and lowers the device's energy consumption up to $60.4$\%.
These findings demonstrate the potential of HI and EE-HI to enable more efficient ML in IoT systems.
%on Jetson Orin for CIFAR-10 image classification.
%Therefore, we design EE-HI systems by incorporating the early exit technique and  Accepting high-confidence inferences on early-exit branches further improves the responsiveness and lowers the device's energy consumption of the HI system in some scenarios.
\end{abstract}


\IEEEpeerreviewmaketitle

\section{Introduction}\label{sec:intro}
%Importance of on-device inference
%On-device ML inference holds great promise for real-time decision-making, enabling applications like image recognition, natural language processing, and predictive analytics to be run directly on the device. 
Deep Learning (DL) inference on data generated by \textit{end devices} such as IoT sensors, smartphones, and drones enables significant improvements in operational efficiency and functionality across various sectors, including industrial automation, smart cities, remote healthcare, smart agriculture, and other sectors. Compared to offloading to the cloud, performing DL inference locally on an IoT device has the added advantage of increased energy efficiency \cite{banbury2021micronets}, responsiveness, and privacy and holds great promise for real-time decision-making. Thus, significant efforts have been directed towards designing and developing compact DL models tailored explicitly for deployment on end devices \cite{deng2020model,chen2020deep,sanchez2020tinyml}.
%There is a significant research thrust in building tinyML models to enable complex tasks such as image classification for medical diagnosis, natural language processing for chatbots, and predictive analytics for fraud detection. 
This resulted in a plethora of DL models, such as MobileNet \cite{sandler2018mobilenetv2}, EfficientNet \cite{tan2019efficientnet}, and Gemma 2B large language model \cite{gemmateam2024}, which can be embedded on moderately powerful smartphones. TinyML models like ResNet-8 \cite{banbury2021mlperf} have also been designed for extremely resource-limited IoT sensors with micro-controller units (MCUs). 

\begin{figure} [t!] 
    \begin{center}
  \includegraphics[width=8.5cm]{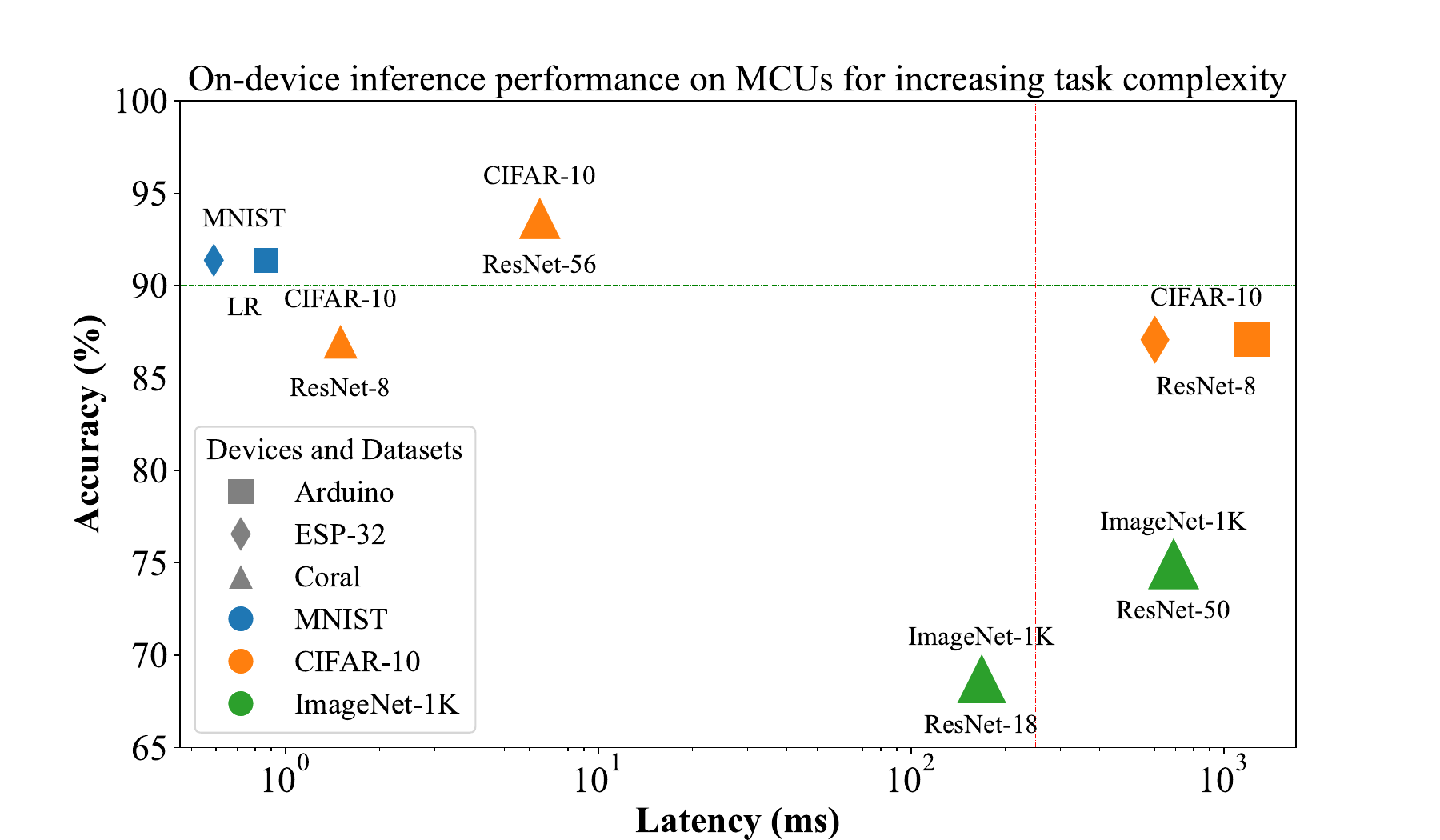}
    \end{center}
    \caption{Measured inference accuracy and latency per inference for various models on different MCUs: Arduino Nano, ESP-32, and Coral Micro, for the MNIST, CIFAR-10, and ImageNet-1K datasets. As the inference complexity increases from MNIST to ImageNet-1K, even resorting to larger DL models and powerful MCU Coral Micro (with TPU) may be insufficient to satisfy QoS requirements: $90$\% accuracy (green line) and $250$ ms response time (red line).} \label{fig:PC_MCUs}
\end{figure}

%Especially when considering DL inference on MCUs, 
Despite the advances in ML model optimization, the large-scale adoption of on-device inference is limited to applications involving simple inference tasks such as visual keyword spotting on smartphones, gesture recognition on smart cameras, and predictive analytics for industrial machines on IoT sensors \cite{imagimob}. 
%Given this, achieving an advantageous trade-off between accurate inference predictions and inference speed often necessitates more complex models, which are challenging to deploy on resource-limited devices. 
We indeed argue that on-device inference becomes inadequate as the complexity of the inference tasks increases. For this purpose, consider an image classification application\footnote{We focus on image classification datasets and models as they span a broad range of complexity and are widely used benchmarks for evaluating the performance of DL systems \cite{reddi2020mlperf}.} with a QoS requirement of at least $90$\% accuracy and maximum $250$ ms latency\footnote{This value is motivated by the latency requirement of $100$ to $500$ ms in real-time computer vision applications such as scene text detection \cite{zhou2017east}, and image captioning \cite{zhou2021semi}.}.
%-- where QoS entails the entire duration required for an inference task to be executed by an end device while ensuring an average accuracy above a certain threshold and within a time limit. 
In Fig. \ref{fig:PC_MCUs}, we present the accuracy and latency for different ML models on three MCUs: Arduino Nano, ESP-32, and Coral Micro (cf. Table \ref{tab:mcus}). Of the MCUs, Coral Micro is the most powerful, featuring a Tensor Processing Unit (TPU). For the inference tasks, we categorize MNIST image classification as a \textit{simple task}, CIFAR-10 as a \textit{moderately complex task}, and ImageNet-1K as a \textit{complex task}. Observe that Arduino Nano and ESP-32 satisfy the QoS requirement for the MNIST dataset using Logistic Regression (LR). But both devices fall short of latency and accuracy requirements for CIFAR-10 dataset because running the state-of-the-art tinyML model ResNet-8, which has $87$\% accuracy, takes at least $600$ ms. Coral Micro can meet the QoS requirements for CIFAR-10, using a bigger DL model ResNet-56, but it cannot do so for ImageNet-1K. Thus, resorting to more computationally capable end devices, in the microcontroller space, may still be inadequate to support emerging ML applications with ever-increasing computation demands.

An alternative solution for network-connected devices is to leverage an edge server or cloud equipped with a state-of-the-art, large-scale deep learning model.
%Today, over $15$ billion devices are connected to the network, which is predicted to increase significantly in 6G networks. 
%end devices are being increasingly exposed to good-quality networks. 
By offloading all the data samples for \textit{remote inference} on the large-size DL model, a device can achieve higher accuracy rather than inferring locally. However, fully offloading the execution of the DL inference risks: \emph{(i)} underutilizing the on-device capabilities, \emph{(ii)} introducing additional network latency and connectivity dependency, and \emph{(iii)} forfeiting the energy efficiency and responsiveness benefits of the on-device inference. Consequently, extensive research has lately focused on distributed inference techniques that use both on-device and remote-server computing capacity. These techniques include \emph{(i)} DNN-partitioning \cite{Kang2017}, \emph{(ii)} inference load balancing \cite{Fresa2023}, and \emph{(iii)} Hierarchical Inference (HI) \cite{Moothedath2024}. DNN-partitioning computes inference using a large-size DL model by partitioning the model's layers between the device and a remote server. Nevertheless, the benefits of this technique have only been realized for computationally powerful devices (such as smartphones) with mobile GPUs \cite{jointoptimization,ebrahimi2022combining}. Furthermore, the technique is not feasible for proprietary DL models, as the model architecture may not be openly available. The inference load-balancing approach divides the computational load between the device and the server to minimize inference time or energy consumption \cite{Ogden2020,Nikoloska2021,Fresa2023}.

\begin{figure*} [ht!] 
    \begin{center}
  \includegraphics[width=0.65\textwidth]{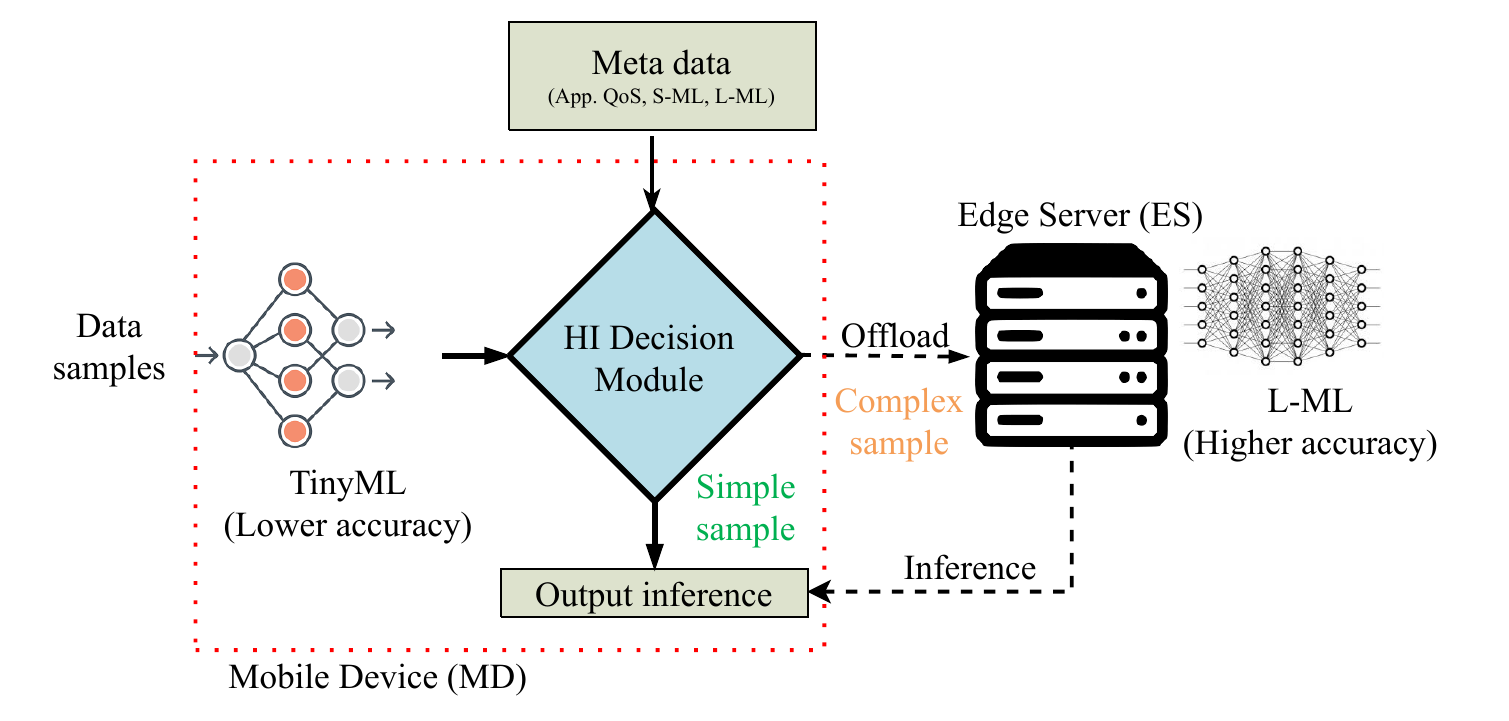}
    \end{center}
    \caption{HI framework for DL inference at the edge.} \label{fig:HI-system}
\end{figure*}

HI \cite{Moothedath2024} paradigm shown in Fig.~\ref{fig:HI-system}, mitigates the limitations of on-device inference by using remote inference only when on-device inference is likely incorrect. A data sample for which the on-device inference is likely correct is a \textit{simple data sample}, and its inference is accepted. In contrast, a data sample deemed likely to result in incorrect local inference is considered a \textit{complex data sample} and is offloaded. HI uses a \textit{HI decision algorithm} that acts on the on-device (local ML) model output for each input sample and decides if the sample is simple or complex.
By only offloading the complex samples, HI reaps the benefits of performing inference on end devices without compromising on accuracy, and thus, it received considerable attention in the recent past~\cite{Nikoloska2021,Ghina2023,behera2023,Moothedath2024,Beytur2024,letsiou2024hierarchical}.

Since HI requires inference on the device for every sample, it introduces a fixed overhead on latency and energy consumption per inference. Additionally, existing work on HI only studied the accuracy improvements over the on-device inference, without accounting for its latency and on-device energy consumption overheads. Furthermore, their conclusions were based only on simulations using abstract costs for offloading and classification errors. Finally, they ignore the diversity and heterogeneity of edge ML systems, including varying hardware, network connectivity, and models. 

In this paper, we address the aforementioned shortcomings and make multiple contributions.
First, we perform accuracy, latency, and energy measurements for running embedded DL models on five devices with different capabilities and three image classification datasets (please refer to Section \ref{sec:system} for additional details about the devices and models used). These measurements will be of independent interest to the embedded ML community as these findings help optimize ML models for low-power, real-time applications, reducing reliance on cloud offloading while maintaining performance. In this respect, we have made the code for device measurements and model training publicly available.\footnote{https://github.com/JoseGaDel/Hierarchical-Inference} Second, we empirically evaluate the performance of HI and on-device inference and demonstrate that HI improves accuracy but has higher latency and energy consumption compared to on-device inference. This raises the question of the true benefits of using HI. Third, in contrast to the previous works, we argue that it is beneficial to choose an HI system that uses a smaller on-device model, even if the model does not meet the accuracy QoS requirement, rather than an HI system with a larger state-of-the-art on-device model that does. Through measurements, we demonstrate that an HI system with a smaller on-device model can achieve the accuracy QoS requirement while delivering lower latency and energy consumption. This highlights the advantages of a carefully designed HI system over standalone on-device inference.
%We demonstrate that a HI system achieves a given accuracy requirement using a small on-device model, even if this model's accuracy is below the specified requirement. Consequently, the HI system using the small model achieves lower latency and energy consumption than a larger on-device model that meets the initial accuracy requirement. This establishes the benefit of using HI over on-device inference.
%Given an accuracy requirement, we demonstrate that HI achieves t using a smaller on-device model (even if its accuracy is below the specified requirement) results in a lower latency and energy consumption than a larger on-device model that meets the initial accuracy requirement.
%a key insight we provide is that, given an accuracy requirement, the latency and energy overheads of a HI system can be reduced by using a smaller on-device model (whose accuracy can be lower than the accuracy requirement). 
For instance, on the Raspberry Pi, with an accuracy requirement of $90$\% for CIFAR-10 image classification, we demonstrate that using a smaller model, ResNet-8, with HI achieves $95.9$\% accuracy while reducing latency by $73$\% and energy consumption by $77$\% compared to the larger on-device model, ResNet-56, which achieves $93.7$\% accuracy. Additionally, HI delivers up to $11$\% (absolute difference) higher accuracy for a given latency and energy requirement for the systems we studied.
Fourth, for ImageNet-1K image classification, we found that offloading to remote inference remains the only viable solution for Coral Micro and Raspberry Pi, as no on-device models can meet the required low latency and energy consumption. 
%An important observation we make is that, in contrast to on-device inference, HI does not require large-size on-device models to meet high accuracy QoS requirements; rather, the model should be small in size, and its output can be effectively used to differentiate between simple and complex samples. This calls for the design of small-size, reasonably accurate embedded ML models whose outputs are better calibrated for differentiating between simple and complex samples.
%Thus, the key to building an efficient HI system is the availability of a smaller-size local ML with reasonable accuracy. 
%This calls for the design of embedded ML models that do not need to compete with state-of-the-art large-size DL models in terms of accuracy but are more efficient regarding their accuracy-to-size ratio. 
To alleviate this issue, we propose using the early exit technique \cite{teerapittayanon2016branchynet} to reduce the latency of inference on the device and consequently further improve the HI approach. In particular, we design a hybrid approach that combines Early Exit with HI (EE-HI) and demonstrate that, compared to HI, EE-HI reduces latency by up to $59.7$\% and device energy consumption by up to $60.4$\%.
%on Jetson Orin for CIFAR-10 image classification.

The rest of the paper is organized as follows. In Section \ref{sec:related}, we discuss the related works and the existing edge ML inference strategies.
In Section \ref{sec:system}, we present the system model along with devices, datasets, models and performance metrics used in this study.
In Section \ref{sec:methodology}, we describe the system methodology used for measurements.
We present the measurement results in Section \ref{sec:measuremet_results}, and compare the performance of HI, on-device, and remote inference in Section \ref{sec:comparison}. We present the design and the results of EE-HI in Section \ref{sec:EEHI} and conclude in Section \ref{sec:conclusion}.

\section{Related Works} \label{sec:related}
\subsection{Existing Strategies for Edge ML Inference}\label{subsec:strategies}
This subsection discusses the related works in edge ML inference systems. The methods that use on-device inference can be broadly classified into 1) on-device inference, 2) DNN partitioning, 3) inference load balancing, and 4) HI. In the following, an \textit{inference task} refers to generating the inference output for a given data sample at the device. For the on-device inference strategy, this task involves running the local DL model on the data sample. For remote ML inference, it entails offloading the data sample to a server and receiving the results. The HI task combines these approaches, either performing standalone on-device inference or using a combination of on-device and remote inference based on the HI offloading decision.

\textit{\textbf{On-Device Inference:}}
As discussed in Section \ref{sec:intro}, considerable effort has been dedicated to the design of compact DL models tailored for deployment on end devices \cite{sanchez2020tinyml,li2024flexnn,zhang2024oncenas,cheng2024advancements}. Further, techniques such as early exit \cite{teerapittayanon2016branchynet} have been explored to reduce the execution time of DL inference. As a result, state-of-the-art compact DL models can perform simple ML inference tasks on devices, ranging from face recognition on mobile phones \cite{lane2017squeezing} to visual wake word and keyword spotting on IoT devices \cite{banbury2021mlperf}. However, the main focus in designing these models has been improving accuracy, while time per inference and energy consumption have not been critically studied.

\textit{\textbf{DNN Partitioning:}}
DNN partitioning, proposed in \cite{Kang2017}, partitions the layers of a large-size DL model and deploys the front layers on the mobile device while the deep layers are deployed on an edge server. It received considerable attention recently \cite{Li2020, Hu2022}. Recent research has even enhanced the performance of DNN partitioning by introducing techniques such as encoding and decoding data transmitted over networks \cite{yao2020deep} and selectively determining which features to process locally versus remotely \cite{huang2022real}. But as demonstrated in \cite{Ghina2023}, for DNN partitioning to be beneficial in reducing the execution time per inference, the processing times of the DL layers on the mobile device should be small relative to the communication time of the data generated between layers. Thus, this technique requires mobile GPUs, making it infeasible for resource-constrained end devices such as IoT devices.

\textit{\textbf{Inference Load Balancing:}} Since the initial proposal of edge computing  \cite{satyanarayanan2009}, significant attention has been devoted to the computational offloading for generic compute-intensive applications \cite{Mach2017}. Due to the growing importance of edge intelligence, recent studies have examined computation offloading for applications using ML inference \cite{wang2018bandwidth,Ogden2020,Fresa2023,hu2021rim}. 
%However, the aspect of accuracy, relevant to offloading data samples, has only received attention recently, as seen in works such as \cite{Nikoloska2021,Fresa2023}. 
This line of research aims to load-balance the inference task by partitioning the data samples between the device and the server, considering parameters such as job execution times, communication times, energy consumption, and average test accuracy of ML models. Nonetheless, in \cite{wu2022distributed}, the authors demonstrated that load-balancing data samples may lead to adversarial scenarios where most data samples scheduled on the device receive incorrect inferences. HI circumvents this issue by utilizing the local DL output and offloading a data sample if its local inference is likely incorrect.

%\subsubsection*{\textbf{Remote Inference}}
\textit{\textbf{Hierarchical Inference:}}
%\JP{Elaborate on softmax values and why they are used for differentiating simple and complex samples.}
The key element is the HI decision algorithm for differentiating simple and complex samples, and so driving the offloading decision. To this end, all existing works use the soft-max values (confidence values) output by the local DL model corresponding to each class. An image is classified into the class that has the maximum soft-max value. Nikoloska and Zlatanov \cite{Nikoloska2021} computed the HI offloading decision by computing a threshold for the maximum soft-max value based on the transmission energy constraint of the device. Al-Atat et al. \cite{Ghina2023} proposed a general definition for HI, presented multiple use cases, and compared HI with existing distributed DL inference approaches at the edge. Similar to \cite{Nikoloska2021}, a threshold was computed based on the trade-off between local misclassifications and offloading costs. Behera et al. \cite{behera2023} showed that using a binary LR on the first two highest soft-max values can improve the HI offloading decision. The binary LR is trained to learn which images are correctly classified and which are incorrectly classified by the local DL. Letsiou et al. \cite{letsiou2024hierarchical} proposed a batched HI approach that offloads samples in batches to minimize communication overhead and enhance system efficiency while maintaining performance comparable to existing HI approaches. In this paper, we use the binary LR as the HI decision algorithm proposed in \cite{behera2023}. Nonetheless, all the aforementioned works use simulation to evaluate their performance against on-device inference using abstract costs for offloading and misclassification errors.
\vspace{-0.3cm}

\subsection{Existing Measurement-based Studies}
%Edge computing has become a crucial tool for ML inference in resource-constrained environments, with significant progress in understanding and optimizing performance at the edge. 
%In this subsection, we discuss the existing works that present measurement-based studies on different ML inference approaches. 

A key requirement for our work is to measure performance in order to evaluate the efficiency of ML models on different devices. However, measuring ML performance on resource-constrained and heterogeneous devices is challenging, as different hardware platforms typically require different software packages to run the models. To address this challenge, the ML performance (MLPerf) benchmarking organization has recently proposed a benchmarking suite, called TinyMLPerf, for tinyML models to perform simple inference tasks \cite{banbury2021mlperf}. Throughput and latency measurements were presented in \cite{torelli2021measuring} for ResNet-50 and MobileNet on Jetson Nano and Coral Micro. 
%The suite comprises of TinyML tasks common to usual use-cases. 
%More fine-grained and precise profiling on Android devices has been proposed by Chu et al. \cite{chu2024}, who present \texttt{nnPerf}, a profiler for \texttt{TFLite} models. However, none of the above works provide energy measurements or compare edge ML strategies. 
Chu et al. \cite{chu2024} have proposed a more detailed and precise profiling method for Android devices through their tool, \texttt{nnPerf}, explicitly designed for \texttt{TFLite} models. Still, none of the above studies offer energy measurements or compare edge machine learning strategies.

An empirical comparison of on-device and cloud-based inference is presented in \cite{guo2018cloud}, illustrating the latency, energy, and accuracy trade-offs between the two approaches. The comparative study was limited to centralized techniques, leaving distributed inference mechanisms unexplored. Furthermore, some studies bench-marked diverse edge devices, such as the NVIDIA Jetson Nano (``end-of-life" as of March 2022) and Google Coral Dev Board, analyzing the on-device inference time \cite{tobiasz2023edge}, power consumption, and accuracy under varying workloads and model configurations \cite{baller2021deepedgebench}.
%\cite{baller2021deepedgebench} - In this paper,they studied MobileNet and used different methodology for energy measurement. In their methodology, they collected the power measurement on a monitoring device and use Network Timing Protocol to synchronize the time between the device and the monitor.
Similarly to \cite{guo2018cloud}, these studies emphasized hardware-software performance comparisons without exploring the impact of distributed inference strategies.
%The embedded ML community has created the open source project MLPerf Tiny benchmark\footnote{https://github.com/mlcommons/tiny/tree/v1.1}, which provides a representative set of deep neural nets and benchmarking code to compare performance between embedded devices. 
The authors in \cite{liang2020ai}, conducted a measurement-based study exploring architectural trade-offs in deploying DNNs on edge devices with AI accelerators, focusing on model compression and DNN partitioning highlighting how these techniques balance energy efficiency, computational demands, and system performance, enabling edge devices with AI accelerators to support AI workloads effectively. However, as highlighted in \cite{li2019edge}, DNN partitioning is ineffective on devices lacking hardware accelerators, such as Raspberry Pi or MCUs. In contrast, this work presents a comprehensive measurement-based study evaluating the energy, latency, and accuracy trade-offs across on-device, remote, and HI approaches, covering a diverse range of edge devices from smaller MCUs to more capable platforms like the Jetson Orin. Furthermore, we also propose the EE-HI approach that reduces the latency and energy consumption of an HI system.
%\vspace{-0.4cm}
\subsection{Early Exit with various ML Inference}
%\subsection{Early Exit with Hierarchical Inference}
%\JP{This is not an existing strategy. Not sure where to place it.}
%\textcolor{red}{Along with the strategies mentioned, we propose to extend the HI state-of-the-art by introducing a new strategy that complements HI with an early exit technique.}
Early exit techniques \cite{rahmath2024early, laskaridis2021adaptive, jazbec2024fast} in DNNs optimize inference by enabling predictions at intermediate layers, thereby reducing computation and latency. BranchyNet \cite{teerapittayanon2016branchynet} pioneered this approach, introducing confidence-based thresholds for early classification, specifically tailored for resource-constrained environments. MSDNet \cite{huang2017multi} advanced the concept by combining multi-scale feature extraction and early exits, allowing simpler inputs to exit early while processing more complex ones through deeper layers. EENet \cite{demir2019early} introduced adaptive thresholds for dynamic, context-aware exits, while PTEENet \cite{lahiany2022pteenet} augmented pre-trained models with task-prioritized early exits for improved resource allocation. In \cite{zhang2021self}, early exits are enhanced by self-distillation, guiding learning with intermediate layers to preserve accuracy. Innovations like EdgeExit \cite{wang2023tlee} tailored early exits for heterogeneous edge systems, ensuring low-latency, power-efficient inference, while reinforcement learning-based methods \cite{nordaunet2023competitive} improved exit decisions through dynamic policy learning. These techniques are widely used in end devices for efficient, on-device inference.

Early exit has also been integrated with DNN partitioning methods to further enhance inference. In \cite{colocrese2024early}, DNNs are partitioned for efficient edge computing with early exits, optimizing resource usage on edge devices. Conversely, \cite{laskaridis2020spinn} proposes a hybrid method, partitioning DNNs across edge devices and the cloud, dynamically offloading computation as needed. Similar approaches in \cite{ebrahimi2022combining, li2021dnn, kim2024early} demonstrate the effectiveness of combining early exits with DNN partitioning to improve inference efficiency. In contrast to previous works, we design a novel EE-HI system that uses the early exit technique to further reduce the latency and energy consumption of the HI system. More details of this method are presented in Section \ref{sec:EEHI}.

\section{ML Models and Performance Metrics}\label{sec:system}
\subsection{Devices}
\label{sec:devices}

\begin{figure*} [ht]
    \begin{center}
    \includegraphics[width=16.5cm]{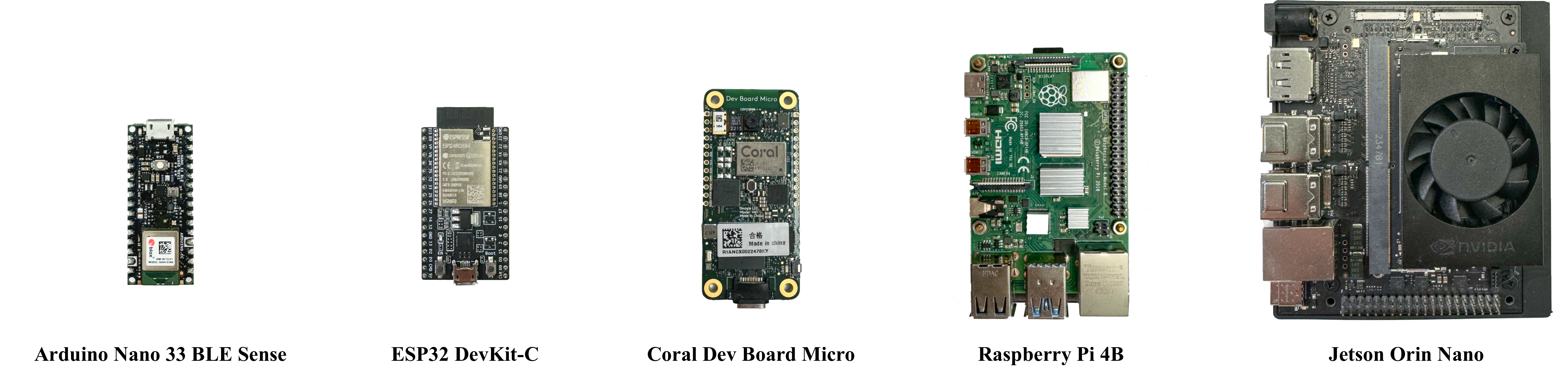}
    \end{center}
    \caption{Devices used in the study. Starting from the left: Arduino Nano 33 BLE Sense, ESP32 DevKit-C, Coral Dev Board Micro, Raspberry Pi 4B, Jetson Orin Nano.} \label{fig:devices}
\end{figure*}

In this study, we conduct measurements on five devices shown in Fig. \ref{fig:devices} and described in Tables \ref{tab:mcus} and \ref{tab:moderately-powerful-devices}. 

\subsubsection*{\textbf{Arduino Nano and ESP32}}\label{sec:nano33}
Arduino Nano 33 BLE Sense and ESP32 (cf. Table \ref{tab:mcus}) are the least powerful devices used in this paper. On-device inference is performed using the TensorFlow Lite Micro framework, which requires loading the model as a binary file, leading to slower on-device computation. 
Furthermore, in contrast to all other devices, Arduino Nano does not support Wi-fi communication. For this reason, offloaded images are transmitted to the edge server via Bluetooth Low Energy (BLE).

\subsubsection*{\textbf{Coral Micro}}
\label{sec:coral-micro}
Coral Dev Board Micro (cf. Table \ref{tab:mcus}) posseses higher processing power and more memory than other commodity MCUs. It can load and execute large compute-intensive DL models with higher inference accuracy. Although the device has a built-in TPU, in certain cases, it falls short. For example, the TPU supports a limited set of instructions\footnote{https://coral.ai/docs/edgetpu/models-intro/\#supported-operations}, which directly impact the models it can execute using the TPU. The \texttt{TFLite} models must be compiled using \texttt{edgetpu-compiler} to be executed on the TPU. Further, because EE models use an intermediate layer with an \texttt{if} instruction, they cannot be compiled to a model supported by the TPU because the instruction is not supported. One potential mitigation of such an issue is running the model on the CPU at the expense of increased inference latency and energy consumption.

\subsubsection*{\textbf{Raspberry Pi 4B}}
Contrary to Jetson Orin or Coral Micro, the Raspberry Pi 4B
(cf. Table \ref{tab:moderately-powerful-devices}) does not benefit from any hardware-based AI acceleration. Such selection provides us with a broader view of the available system-on-chip ecosystem. Regarding the software implementation, all the models are loaded and executed using the \texttt{TFLite} Python API. In this framework, all the necessary operations for the tested models are supported\footnote{https://www.tensorflow.org/mlir/tfl\_ops}. Combined with the fact that the models were run on the CPU, this enables smooth deployment, even for Early Exit implementations.

\subsubsection*{\textbf{Jetson Orin}}
Jetson Orin Nano (cf. Table \ref{tab:moderately-powerful-devices}) includes an integrated GPU that features Tensor Cores.
%They are programmable fused matrix-multiply-and-accumulate units that execute concurrently alongside the CUDA cores and support half-precision and integer instructions. This are common operations in CNNs, which make them ideal for computer vision workloads. 
For this device, we have used TensorRT SDK\footnote{https://docs.nvidia.com/deeplearning/tensorrt}, which is a framework developed by NVIDIA for high-performance deep learning inference, built on top of CUDA. It optimizes the models using layer and tensor fusion and kernel tuning to maximize performance on the specific hardware of the target device. 
To port the pre-trained models to TensorRT, we must first convert them to ONNX.
This pipeline yields correct results for base models, but for early exit we find that the TensorRT engine fails to produce a graph-capturable network in CUDA, which causes early exit models to be slower than the base implementations. To mitigate this issue, transforming the Tensorflow subgraph models to PyTorch successfully generates models that effectively utilize CUDA graphs and achieve the required performance.
% The Jetson Orin provides power and performance management features, allowing users to tweak the device to their needs. 
%For example, the \texttt{cpufreq} subsystem allows for dynamic voltage and frequency scaling (DVFS), where clock rate and voltage are adjusted in real time to instant load. 
Jetson Orin has two power modes: one for maximum performance (15W) with all hardware and clocks at full capacity and another for low power consumption (7W) with RAM, CPU, GPU, and other components capped. We examine both power modes to investigate their effects and observe that enabling dynamic clock rate results in higher latency and variability for relatively small energy savings. Setting the clock to maximum in each power mode results in higher performance per watt and more stable outcomes under constant load.

\subsection{Datasets, Models and Training}
\label{sec:models_and_training}
We study image classification tasks using three datasets MNIST, CIFAR-10, and ImageNet-1K. As discussed in Section \ref{sec:intro}, MNIST image classification tasks can be performed on Aruduino Nano and ESP32, without offloading to a remote ML. Thus, we focus on CIFAR-10 and ImageNet-1K image classification tasks in the rest of the paper. The details of the datasets are presented in Table \ref{tab:datasets}.

\begin{table*}[ht]
\centering
\caption{Device characteristics of MCUs used for the measurements}
\label{tab:mcus}
\begin{tabular}{p{2cm}|p{4.7cm}|p{4.7cm}|p{4.7cm}|}
\cline{2-4} & \textbf{ESP32 DevKit-C} & \textbf{Arduino Nano 33 BLE Sense} & \textbf{Coral Dev Board Micro} \\ \hline
\multicolumn{1}{|l|}{\textbf{CPU}} & Tensilica Xtensa LX6 Dual-Core @ 240 MHz, 160 MHz or 80 MHz & Nordic nRF52840 ARM Cortex-M4 CPU @ 64 MHz & ARMv7 Cortex-M7 @ 800 MHz / M4 @ 400 MHz \\ \hline
\multicolumn{1}{|l|}{\textbf{Memory}} & 520KB SRAM + 8 KB RTC FAST + 8 KB RTC SLOW & 256KB SRAM & 64MB RAM \\ \hline
\multicolumn{1}{|l|}{\textbf{Accelerator}} & - & - & Coral Edge TPU \\ \hline
\multicolumn{1}{|l|}{\textbf{Connectivity}} & Wi-Fi 802.11 b/g/n, Bluetooth v4.2 BR/EDR and BLE & 5.0 BLE & Wi-Fi 802.11 a/b/g/n/ac, Bluetooth 5.0 %(requires an additional
add-on board) \\ \hline
%\multicolumn{1}{|l|}{\textbf{Power Supply}} & 5V & 5V & 5V \\ \hline
\end{tabular}
\end{table*}

\begin{table*}[ht]
\centering
\caption{Device characteristics of the moderately powerful devices used for the measurements}
\label{tab:moderately-powerful-devices}
\begin{tabular}{p{2cm}|p{7cm}|p{7cm}|}
\cline{2-3}
& \textbf{Jetson Orin Nano}& \textbf{Raspberry Pi 4B} \\ \hline
\multicolumn{1}{|l|}{\textbf{CPU}} & ARMv8.2 Cortex-A78AE @ 1.5 GHz & ARMv8 Cortex-A72 @ 1.5GHz \\ \hline
\multicolumn{1}{|l|}{\textbf{Memory}} & 8GB 128-bit LPDDR5-2133 & 8GB LPDDR4-3200 \\ \hline

\multicolumn{1}{|l|}{\textbf{Accelerator}} & Ampere GPU: 1024 CUDA cores + 32 Tensor cores & - \\ \hline
\multicolumn{1}{|l|}{\textbf{Connectivity}} & Gigabit Ethernet, 2.4GHz and 5GHz IEEE 802.11.b/g/n/ac wireless, Bluetooth 5.0 & Gigabit Ethernet, 802.11ac Wi-Fi, Bluetooth 5.0\\ \hline
%\multicolumn{1}{|l|}{\textbf{Power Supply}} & NSW26493, PA-1450-26 : 19.0V - 2.37A & KSA-15E-051300HE : 5.1V - 3.0A \\ \hline
\end{tabular}
\end{table*}

%\subsection{Base Models and Training} 

We use TensorFlow\footnote{https://www.tensorflow.org/} (specifically, TFLite\footnote{https://www.tensorflow.org/lite} and TFLite Micro\footnote{https://www.tensorflow.org/lite/microcontrollers}) as the DL software library for most of this work due to its popularity and convenience in compiling models for on-device deployment. We refer to the models we introduce below as \textit{base models} to differentiate them from the models we design and train using early exit branches, later in Section \ref{sec:EEHI}.
\\
%\subsubsection{CIFAR-10:}
\textbf{CIFAR-10.} Three models are selected for this dataset. The smallest one being an implementation of ResNet-8 presented by MLPerf\footnote{https://github.com/mlcommons/tiny}. Another model from the ResNet \cite{he2016deep} family is selected as an alternative: ResNet-56. In this case, we trained the model, achieving an accuracy of up to 93.77$\%$. Due to its popularity, AlexNet \cite{krizhevsky2012imagenet} is also considered in this work. We trained the model up to 74.51$\%$ accuracy. 
\\
\textbf{ImageNet-1K.} For ImageNet, we select ResNet-18, ResNet-50, AlexNet, and RegNetY32GF \cite{radosavovic2020designing}. All these models are retrieved from PyTorch's own model repository\footnote{https://pytorch.org/vision/stable/models.html}.  Nonetheless, in an attempt to facilitate deployment and preserve framework uniformity, these models are converted to TensorFlow, maintaining the same structure and accuracy.
\\
\textbf{Early Exit (EE) models.}
We design on-device early exit models for CIFAR-10 models (ResNet-8, ResNet-56, AlexNet) by implementing the methodology presented in \cite{teerapittayanon2016branchynet}. Our approach differs from \cite{teerapittayanon2016branchynet} in that we use a threshold on the model confidence, i.e., the maximum softmax value, instead of the entropy of all the softmax values. Later in Section \ref{sec:EEHI}, we present the details of our design of the early exit models.

\begin{table*}[ht]
\caption{Details of the datasets used}
    \label{tab:datasets}
    \centering
    \begin{tabular}{|l|c|c|c|}
        \hline
        \textbf{Dataset}      & \textbf{Number of Samples} & \textbf{Input Size} & \textbf{Number of Classes} \\
        \hline
        \textbf{MNIST}       & 70,000 (60,000 train, 10,000 test) & $28 \times 28$ grayscale & 10 \\
        \textbf{CIFAR-10}    & 60,000 (50,000 train, 10,000 test) & $32 \times 32 \times 3$ RGB & 10 \\
        \textbf{ImageNet-1K} & $\sim$1.28M (train), 50,000 (val), 100,000 (test) & $224 \times 224 \times 3$ RGB (varies) & 1,000 \\
        \hline
    \end{tabular}
\end{table*}

\subsection{Performance Metrics and QoS}\label{subsec:QoS}
We measure the performance of on-device inference, HI, EE-HI, and remote inference strategies using the following three key metrics commonly employed in QoS requirements in edge ML applications\footnote{https://mlcommons.org/benchmarks/inference-tiny/}. Each metric is defined as the average over the set of data samples from a dataset.
\begin{itemize}
    \item \textbf{Energy per Inference (EPI):} It refers to the average energy consumed by the device during an inference task\footnote{We only consider the on-device energy consumption, taking into account that devices can be battery limited.}. 
    %Minimizing energy consumption is vital for extending the battery life of end devices and ensuring sustainable operation, especially in resource-constrained scenarios.
    \item \textbf{Latency per Inference (LPI):} It is the average time taken to process an inference task. 
    %Lower latency is essential for applications that require real-time decision making.
    \item \textbf{Accuracy:} It is the percentage of correct inferences (top-1 accuracy) provided for the validation dataset. 
    %Maintaining high accuracy is critical to ensure the reliability and effectiveness of edge AI applications.
    %https://mlcommons.org/benchmarks/inference-tiny/
\end{itemize}

Given different datasets with varying inference task complexity, choosing QoS metrics is non-trivial because the range of accuracy values achieved in edge ML systems is limited by the availability of on-device and remote DL models, which depend on the datasets. Given this difficulty, we choose the following QoS requirements with some justifications.
%Therefore, we define the following QoS requirements with some justifications.

%Given the breadth of edge ML applications and their varying requirements, choosing appropriate QoS metrics is challenging. While preceding works used specific values for the above metrics as the QoS requirements, we chose them as parameters dependent on the remote inference for the reasons stated below
%For these metrics, we define the following three QoS requirements. 

\textit{Accuracy QoS constraint:}  Motivated by \cite{wu2023qos}, we choose the accuracy requirements $10$\% lower than the state-of-the-art (SOTA) model accuracy for the respective datasets, $99.8$\% for MNIST \cite{mazzia2021efficient}, $99.5$\% for CIFAR-10 \cite{dosovitskiyimage}, and $90$\% for ImageNet-1K \cite{dehghani2023scaling}). Thus, the accuracy QoS requirement for MNIST and CIFAR-10 image classification is to achieve at $90$\%, and for ImageNet-1K is to achieve at least $80$\%.
%The accuracy QoS requirement is to achieve at least $0.9\times$ the highest accuracy achieved by any state-of-the-art model for the given dataset. The highest accuracy models available for MNIST, CIFAR-10, and ImageNet-1K achieve $99.8$\% \cite{mazzia2021efficient}, $98.77$\% \cite{dosovitskiyimage}, and $90$\% \cite{dehghani2023scaling}), respectively. This translates to the QoS greater than $90$\% accuracy requirement for MNIST, greater than $89$\% for CIFAR-10, and greater than $81$\% for ImageNet-1K. We note that choosing a higher accuracy than the above QoS requirement makes on-device inference infeasible, as the devices under study cannot support large-size DL models required to achieve such high accuracies.

%This choice was guided by the fact that, for the datasets under consideration, the size of the DL models that achieve higher accuracy than $0.9\times$ the highest accuracy increases exponentially \cite{tan2019efficientnet}, making their deployment on the devices either infeasible or inferior to remote inference. 

\textit{Latency and energy constraints:} 
The \textit{average latency QoS} requirement in this context is defined dynamically, requiring inference to be completed in less than half the time needed to offload an image and receive the result back at the device. Similarly, the \textit{average energy QoS} requirement is dynamically set to use less than half the energy required for the same offloading process.
%The latency (energy) QoS requirement is to achieve less than half the time required (less than half the energy consumed on the device) to offload an image and receive the inference back at the device. 

%\textit{Maximum energy constraint:} The energy QoS requirement is to achieve less energy consumption than half the energy consumed on the device to offload an image and receive the inference back at the device. 

Note that if latency or energy constraints are relaxed enough to permit offloading an image, remote inference is the best strategy as it achieves higher accuracy. When comparing different strategies in Section \ref{sec:comparison}, we fix one of the QoS requirements and determine the strategy that provides the maximum gains for the other two metrics.

%Achieving this QoS is the primary goal. However, one needs to understand that achieving the primary goal is a necessary condition, but not sufficient. When we talk about different strategies achieving the QoS, it refers to achieving that QoS as a constraint while optimizing the other available resources. So, we can redefine our QoS as "minimizing time and energy (as they are proportional) subject to constraint that accuracy is at least 90\% of SOTA", and similarly, "maximizing accuracy subject to constraint time per inference is less than 50\% of offloading time" and "maximizing accuracy subject to constraint energy per inference is less than 50\% of offloading energy". We consider these specific requirements to have a fair comparison with remote inference.

\section{Methodology for Measurements}\label{sec:methodology}
%\subsection{Methodology for Measurements}
\begin{comment}
The three aspects are further divided into two separate measurement categories: \textit{Device} and \textit{System}. The \textit{Device} category represents measurements taken on device (e.g., inference accuracy performed solely on device), while \textit{System} provides a more holistic view of the HI-based system providing metrics of the entire inference pipeline involving both the device and the more capable machine to which the image data is offloaded for inference. 

For these measurements, a fully integer quantized version of the models was used. The inference workflow in the proposed HI system is the following: 
\begin{enumerate}
    \item Retrieve the image and preprocess it to the desired format.
    \item Run inference on the device.
    \item Select the highest confidence values and perform Logistic Regression (LR).
    \item If the output of LR is favorable, accept the predicted class by the device. Otherwise, proceed to offload the image.
    \item In case the image is set to be offloaded, send it via the network to the edge server.
    \item Once the image is fully received on the server, preprocess it to the accepted format of the server model.
    \item Run inference on the server.
    \item Send the predicted class by the server model back to the device via the network.
    
\end{enumerate}
\end{comment}

\subsubsection{On-device} All the models examined in this work are either pre-trained by the authors or obtained from existing libraries. Given that \texttt{INT8} quantized models have lower latency and energy consumption \cite{hubara2018quantized}, we build and use quantized models albeit at the expense of slightly lower accuracy than the full-precision models. For each device, we load a pre-trained model and measure its latency and energy consumption by performing inference on $n$ images, where $n$ is the size of the validation dataset. The averages are then computed, with $n$ varying according to the size of the validation dataset. Note that, for the inference evaluation, we measure the latency and energy consumption of the API call that initiates the inference routine in each framework. % This includes preprocessing the image into the desired dimension required by the model.

The power consumption measurements are performed through a Voltech PM1000+ Power Analyzer and its Universal Breakout Box, which allows devices to be plugged directly into a socket to obtain results. Since the Voltech PM1000+ relies on legacy software, we created a custom script to issue remote commands to the power analyzer using the \texttt{PyVISA}\footnote{https://pyvisa.readthedocs.io/en/latest/} library. We obtain the energy values from the average power consumption and the time taken by the underlying inference time or offloading time measurements for inference time and energy consumption for each model. The measurements have a negligible standard deviation. For all the measurements reported in this section, the maximum standard deviation occurred for the power measurement of running AlexNet on Coral Micro, and its value is $6.1$\% of the mean value.
%The energy and latency measurement for the HI system is more involved due to the offloading aspect. 
\subsubsection{HI} For the HI system, we choose ResNet-8 and ResNet-18 as the on-device DL models for CIFAR-10 and ImageNet-1K datasets, as both are tinyML models with a reasonable accuracy-to-size ratio. 

\textit{The binary LR algorithm:} For the HI decision-making, we use a binary LR algorithm for each dataset. We train the LR using the soft-max values output by the on-device model for each image (from the training set) as the input. The output labels during the training are ${0,1}$, where $0$ corresponds to on-device inference being incorrect and $1$ corresponds to the inference being correct.

Specifically, in the testing phase, for the $i$-th image, the inference is accepted if the binary LR outputs $1$, denoted by $LR_i = 1$; otherwise, the image is offloaded for remote inference, denoted by $LR_i = 0$. If the image is set to be offloaded, this is transmitted via the local WiFi network (or Bluetooth in case of Arduino Nano), to the edge server. Once the image is received on the server, it is preprocessed to the format required at the server model. The server processes the prediction and sends the predicted class back to the device via the network. To quantify the accuracy of the HI system, we define key terms: let $dev\_inf_{i}$ and $ser\_inf_{i}$ denote the device and server predictions, and
$gt_{i}$ denotes the ground truth class for image $i$. The HI system's accuracy, denoted by $Acc_{HI}$, is given by,  
\begin{align}
%\tiny
&Acc_{HI} \!\!=\!\! \frac{1}{n}\sum_{i = 1}^{n}\!\Big[\! \mathbbm{1}\!(LR_{i} \!=\! 1) \mathbbm{1}\!(dev\_inf_{i} \!=\! gt_{i})  \nonumber\\
& \quad\quad\quad\quad\quad\quad+  \mathbbm{1}\!(LR_{i} \!=\! 0) \mathbbm{1}\!(ser\_inf_{i} \!=\! gt_{i})\Big] \label{accuracy_eq}\\
&= \!\!\frac{1}{n}\sum_{i = 1}^{n}\! \Big[\!  \mathbbm{1}\!(dev\_inf_{i} \!=\! gt_{i}) \! - \! \mathbbm{1}\!(LR_{i} \!=\! 0) \mathbbm{1}\!(dev\_inf_{i} \!=\! gt_{i})\nonumber \\
& \quad\quad\quad\quad\quad\quad+  \mathbbm{1}\!(LR_{i} \!=\! 0) \mathbbm{1}\!(ser\_inf_{i} \!=\! gt_{i})\Big] \nonumber \\
&=\! Acc_{OD} \!-\! \eta_{FN} \! + \!\frac{1}{n}\!\sum_{i = 1}^{n}\! \! \mathbbm{1}\!(LR_{i} \!=\! 0) \mathbbm{1}\!(ser\_inf_{i} \!=\! gt_{i}), \label{acuracy_alt_eq}
\end{align}
% Alternatively the accuracy of the HI system can also be represented as 
% \begin{equation}
%     Acc_{HI} = (Acc_{OD} - \eta_{FN}) + (Acc_{S} * \eta_{off})
%     \label{acuracy_alt_eq}
% \end{equation}
%where the accuracy obtained by the on-device and remote inference are denoted as $Acc_{OD}$ and $Acc_{S}$ respectively. $\eta_{FN}$ and $\eta_{off}$ represent the fraction of false negative (FN) samples from the HI offloading algorithm and the fraction of offloaded samples.
where $\mathbbm{1}(\cdot)$ is an indicator function, $Acc_{OD}$ represents the accuracy of the on-device model, and $\eta_{FN}$ denotes the fraction of false negatives under the binary LR, i.e., the fraction of data samples offloaded by the binary LR that are correctly classified by the on-device model.
From \eqref{acuracy_alt_eq}, we infer that while the on-device model's accuracy may fall below the accuracy QoS requirement, $Acc_{HI}$ can still meet the requirement if the remote model's accuracy is significantly higher than the QoS requirement and $\eta_{FN}$ is sufficiently small (ideally zero). This underscores the critical role of the binary LR's performance in HI decision-making. From the above observations, we conclude that employing an HI system with a smaller on-device model, even with lower accuracy, can lead to improved overall performance. It is worth noting, however, that a naive HI decision-making algorithm could reduce $\eta_{FN}$ by increasing the number of offloads, thereby boosting overall system accuracy $Acc_{HI}$. Yet, this approach comes at the cost of increased false positives, as well as higher latency and energy consumption per inference.
%Conversely, the $Acc_{HI}$ decreases as $\eta_{FN}$ increases. This shows the importance of using an efficient HI decision algorithm.

We follow a modular approach for latency and energy measurements for HI. The latency per inference for an image includes the processing time for the on-device model and the binary LR decision, denoted by $t_{dev\_inf}$ and $t_{LR}$, respectively. Let $e_{dev\_inf}$ and $e_{LR}$ denote the respective energy consumptions. If the decision is to offload the image, then the latency accounts for the additional term $t_{off}$, which is the total time taken for image transmission to the server, remote ML inference, and sending back the prediction to the device. Similarly, the additional device energy consumption for an offloaded image is denoted by $e_{off}$, which includes the transmission energy. Let $\eta_{off}$ denote the fraction of offloaded samples in the HI system, then the average time per inference, $Time_{HI}$, and average energy per inference, $Energy_{HI}$, are given by
%Thus, the average latency under HI is given in \ref{latency_eq}, where $\eta_{off}$ is the fraction of offloaded samples. Similarly, the average on-device energy consumption per image is given in \eqref{energy_eq}, where $e_{edge\_off}$ is the energy consumed to obtain the remote inference for an offloaded image.
%total image transmission, server inference (with the corresponding preprocessing), and predicted class transmission to the edge device. 
\begin{align}
Time_{HI} &= t_{dev\_inf} + t_{LR} + \eta_{off} \left[ t_{off}\right]
\label{latency_eq}\\
Energy_{HI} &= e_{dev\_inf} + e_{LR} + \eta_{off} \left[ e_{off}\right]
\label{energy_eq}.
\end{align}

%For energy, we are only concerned with the edge device as it is the part of the system where energy consumption plays a critical role. For specific applications, edge devices may need to rely on batteries. Less energy consuming systems at the edge allow for more autonomy by extending battery life. 

\subsubsection{EE-HI} For HI system with early exit on-device models, the accuracy, latency, and energy can be computed using \eqref{accuracy_eq}, \eqref{latency_eq}, and \eqref{energy_eq}. However, in this case, the average latency $t_{dev\_inf}$ and the energy consumption $e_{dev\_inf}$ of the early exit DL model depend on the chosen threshold(s) $\theta$ at the early exit branch(s). If the inference of an image satisfies the threshold condition at an early exit branch, then $t_{dev\_inf}$ and $e_{dev\_inf}$ will only include the processing till that branch.

\section{Measurement Results}\label{sec:measuremet_results}
%
%\JP{All comparison bar plots here}
\subsection{Server Run times}
We measure the run times of the remote ML inference on two distinct servers with different GPUs: NVIDIA Tesla T4\footnote{https://www.nvidia.com/en-us/data-center/tesla-t4/} and NVIDIA A100\footnote{https://www.nvidia.com/en-us/data-center/a100/}. The comparison of the computing powers of these servers are presented in Table \ref{tab:gpu_comparison}.
For CIFAR-10, we consider a pre-trained vision transformer model ViT-H/14 \cite{dosovitskiyimage}, which we fine-tune to achieve an accuracy of $98.77\%$. For ImageNet-1K, we consider a ConvNexT \cite{liu2022convnet} with $86.58\%$ accuracy. For NVIDIA Tesla T4, we measure an average time per inference of $19.18$ ms for ViT-H/14 (CIFAR-10) and $12.01$ ms for ConvNexT (ImageNet-1K). For NVIDIA A100, the measured times are $4.29$ ms for ViT-H/14 (CIFAR-10) and $4.41$ ms for ConvNext (ImageNet-1K). These numbers are obtained using half-precision (FP16), a format that uses 16 bits to represent floating-point numbers. Both GPUs have specialized hardware for FP16 arithmetic, which increases performance with negligible accuracy loss. Upon arrival, the image needs to undergo pre-processing, including conversion of image dimensions to match the input dimensions of the model. We measure the end-to-end average processing time of an inference request as 6.95 ms for CIFAR-10 and 5.08 ms for ImageNet-1K. The reason for smaller inference time for ImageNet-1k images is that ConvNeXT uses the same dimension image as input as the edge models and does not require converting the image dimensions. Therefore, when the image is received by the server, we can immediately perform preprocessing, which takes an average of 0.7 ms. 
However, preparing data for ViT-H/14 is more involved, as it requires transposing the input buffer to the NCHW format\footnote{https://docs.nvidia.com/deeplearning/cudnn/latest/developer/core-concepts.html} and upscaling to the appropriate dimensions before preprocessing. The additional operations are accountable for the uneven increase in the total execution time compared to ConvNeXT.

Considering that both servers are equipped with NVIDIA hardware, TensorRT is deemed the optimal choice for conducting inference testing. Nonetheless, the size of ViT-H/14 exceeds the 2 GB restriction in the \texttt{Protobuf} (Protocol Buffers) serialized message size\footnote{https://protobuf.dev/programming-guides/encoding/}, rendering it impossible to parse the model into the framework. Instead, the ONNX Runtime was utilized with the CUDA execution provider, given that TensorRT execution provider naturally fails in this framework for the same reason. However, the ConvNeXT model did not have this issue, and the inference was executed using TensorRT execution provider within ONNX Runtime.

Given that the runtimes on A-100 are shorter, we use it to offload the images for HI and remote-inference systems.

\begin{table}[h]
\caption{Computing Resource Comparison of Edge Servers}
    \label{tab:gpu_comparison}
    \centering
    
    \begin{tabular}{|l|c|c|}
        \hline
        \textbf{Feature} & \textbf{NVIDIA Tesla T4} & \textbf{NVIDIA A100} \\
        \hline
        \textbf{Architecture} & Turing & Ampere \\
        \textbf{Memory} & 16GB GDDR6 & 80GB HBM2e \\
        \textbf{Memory Bandwidth} & 320 GB/s & 1,555 GB/s\\
        \textbf{CUDA Cores} & 2,560 & 6,912 \\
        \textbf{Tensor Cores} & 320 & 432 \\
        \hline
    \end{tabular}
\end{table}

% \subsection{Wireless Communication Timings}
% \subsubsection{Image transmission: WiFi vs BLE}
% Given the availability of both BLE and Wi-Fi for all the devices (except Arduino Nano), the question of which one should be used arises. To address this, an experiment was performed on a Raspberry Pi, where both latency and energy were measured for transmitting a CIFAR-10 image.

% As expected, BLE's power consumption is smaller, $2.23$ W, compared to WiFi, which consumes $2.71$ W. However, the transmission time of an image under BLE is $177.13$ ms, two orders of magnitude higher than the $2.65$ ms \color{black} time it takes for WiFi. Thus, the average energy consumption for transmitting an image to the server and receiving two bytes back consumes 7.39 mJ in the case of Wi-fi and 395 mJ in the case of BLE. Given significantly high latency and energy consumption under BLE, we opt for Wi-Fi communication for offloading the images.

\begin{table}[h]
\centering
\caption{Latency (communication + server time) and energy consumption to offload an image using WiFi.}
\resizebox{\columnwidth}{!}{%
\begin{tabular}{cc|c|c|cc|}
\cline{3-6}
 &  & \multirow{2}{*}{\textbf{Coral}} & \multirow{2}{*}{\textbf{RPi}} & \multicolumn{2}{c|}{\textbf{Jetson}} \\ \cline{5-6} 
 &  &  &  & \multicolumn{1}{c|}{\textbf{7W}} & \textbf{15W} \\ \hline
\multicolumn{1}{|c|}{\multirow{2}{*}{\textbf{CIFAR-10}}} & \textbf{Latency (ms)} & 10.37 & 9.68 & \multicolumn{1}{c|}{8.34} & 8.20 \\ \cline{2-6} 
\multicolumn{1}{|c|}{} & \textbf{Energy (mJ)} & 14.61 & 27.67 & \multicolumn{1}{c|}{50.48} & 55.91 \\ \hline
\multicolumn{1}{|c|}{\multirow{2}{*}{\textbf{ImageNet-1K}}} & \textbf{Latency (ms)} & 84.41 & 28.74 & \multicolumn{1}{c|}{16.42} & 16.45 \\ \cline{2-6} 
\multicolumn{1}{|c|}{} & \textbf{Energy (mJ)} & 145.13 & 112.01 & \multicolumn{1}{c|}{111.57} & 124.93 \\ \hline
\end{tabular}
}
\label{off_table}
\end{table}

\begin{table*}[ht!]
\centering
\caption{Accuracy, latency and energy results for Coral Micro, Raspberry Pi and Jetson Orin devices}
\begin{tabular}{ll|lll|lll|}
\cline{3-8}
                                                              &                     & \multicolumn{3}{c|}{\textbf{CIFAR-10}}                                                              & \multicolumn{3}{c|}{\textbf{ImageNet-1K}}                                                            \\ \cline{3-8} 
                                                              &                     & \multicolumn{1}{l|}{\textbf{ResNet-8}} & \multicolumn{1}{l|}{\textbf{ResNet-56}} & \textbf{AlexNet} & \multicolumn{1}{l|}{\textbf{ResNet-18}} & \multicolumn{1}{l|}{\textbf{ResNet-50}} & \textbf{AlexNet} \\ \hline
\multicolumn{1}{|l|}{\multirow{4}{*}{\textbf{Accuracy (\%)}}} & \textbf{Coral}      & \multicolumn{1}{l|}{86.98}             & \multicolumn{1}{l|}{93.66}              & 74.39            & \multicolumn{1}{l|}{68.75}              & \multicolumn{1}{l|}{74.97}              & -                \\ \cline{2-8} 
\multicolumn{1}{|l|}{}                                        & \textbf{Raspberry}  & \multicolumn{1}{l|}{86.97}             & \multicolumn{1}{l|}{93.53}              & 74.27            & \multicolumn{1}{l|}{69.67}              & \multicolumn{1}{l|}{75.86}              & 56.36            \\ \cline{2-8} 
\multicolumn{1}{|l|}{}                                        & \textbf{Jetson 15W} & \multicolumn{1}{l|}{86.91}             & \multicolumn{1}{l|}{93.72}              & 74.74            & \multicolumn{1}{l|}{69.59}              & \multicolumn{1}{l|}{76.01}              & 56.45            \\ \cline{2-8} 
\multicolumn{1}{|l|}{}                                        & \textbf{Jetson 7W}  & \multicolumn{1}{l|}{86.91}             & \multicolumn{1}{l|}{93.72}              & 74.74            & \multicolumn{1}{l|}{69.59}              & \multicolumn{1}{l|}{76.01}              & 56.45            \\ \hline
\multicolumn{1}{|l|}{\multirow{4}{*}{\textbf{Latency (ms)}}}  & \textbf{Coral}      & \multicolumn{1}{l|}{1.50}              & \multicolumn{1}{l|}{6.45}               & 69.41            & \multicolumn{1}{l|}{168.22}             & \multicolumn{1}{l|}{690.43}             & -                \\ \cline{2-8} 
\multicolumn{1}{|l|}{}                                        & \textbf{Raspberry}  & \multicolumn{1}{l|}{1.63}              & \multicolumn{1}{l|}{14.36}              & 4.75             & \multicolumn{1}{l|}{227.66}             & \multicolumn{1}{l|}{461.84}             & 105.04           \\ \cline{2-8} 
\multicolumn{1}{|l|}{}                                        & \textbf{Jetson 15W} & \multicolumn{1}{l|}{0.17}              & \multicolumn{1}{l|}{0.64}               & 0.39             & \multicolumn{1}{l|}{0.84}               & \multicolumn{1}{l|}{1.80}               & 1.48             \\ \cline{2-8} 
\multicolumn{1}{|l|}{}                                        & \textbf{Jetson 7W}  & \multicolumn{1}{l|}{0.25}              & \multicolumn{1}{l|}{1.03}               & 0.63             & \multicolumn{1}{l|}{1.85}               & \multicolumn{1}{l|}{4.04}               & 2.64             \\ \hline
\multicolumn{1}{|l|}{\multirow{4}{*}{\textbf{Energy (mJ)}}}   & \textbf{Coral}      & \multicolumn{1}{l|}{2.14}              & \multicolumn{1}{l|}{9.86}               & 112.00           & \multicolumn{1}{l|}{230.68}             & \multicolumn{1}{l|}{939.69}             & -                \\ \cline{2-8} 
\multicolumn{1}{|l|}{}                                        & \textbf{Raspberry}  & \multicolumn{1}{l|}{6.22}              & \multicolumn{1}{l|}{55.59}              & 19.38            & \multicolumn{1}{l|}{977.33}             & \multicolumn{1}{l|}{2004.21}            & 446.08           \\ \cline{2-8} 
\multicolumn{1}{|l|}{}                                        & \textbf{Jetson 15W} & \multicolumn{1}{l|}{1.38}              & \multicolumn{1}{l|}{5.47}               & 4.43             & \multicolumn{1}{l|}{10.56}              & \multicolumn{1}{l|}{23.60}              & 22.14            \\ \cline{2-8} 
\multicolumn{1}{|l|}{}                                        & \textbf{Jetson 7W}  & \multicolumn{1}{l|}{1.72}              & \multicolumn{1}{l|}{7.12}               & 5.59             & \multicolumn{1}{l|}{16.13}              & \multicolumn{1}{l|}{35.71}              & 27.90            \\ \hline
\end{tabular}
\label{tab:all_results}
\end{table*}

\subsection{Wireless Communication Timings}
\subsubsection{Image transmission -- WiFi vs BLE}
Given the availability of both BLE and Wi-Fi on all devices (except Arduino Nano), a key question arises: which one should be used? To explore this, we conducted an experiment using a Raspberry Pi, measuring both latency and energy consumption while transmitting a CIFAR-10 image to the server and waiting for a response. It is worth mentioning that the energy measurements account for the entire working device, not just the communication module. As expected, BLE's power consumption is smaller, $2.23$ W, compared to WiFi, which consumes $2.71$ W. However, the transmission time of an image under BLE is $177.13$ ms, two orders of magnitude higher than the $2.65$ ms \color{black} time it takes for WiFi. Thus, the average energy consumption for transmitting an image to the server and receiving two bytes back consumes 7.39 mJ in the case of Wi-Fi and 395 mJ in the case of BLE. Given the significantly high latency and energy consumption under BLE, we opt for Wi-Fi communication for offloading the images.

\subsubsection{Measuring offloading time over WiFi}
The total time and device energy consumption of offloading include the transmission of the image over Wi-Fi and receiving the inference back at the device. The device, operating as a client, sends a request to the server, which actively monitors communications. To establish Wi-Fi connectivity, the client connects to a hotspot that provides access to the local network where the server is running. In the experiment, an image is transmitted as a raw byte array over TCP, and the server performs inference using the received array to determine the predicted class, which is then returned as a numerical value. The duration from the start of the offloading process to the reception of the response is measured and averaged over $10,000$ images. Measurements for this process with both datasets are presented in Table \ref{off_table}.

% Transferred results from section 4.

\subsection{Device Results}

The results for base models were acquired on 3 devices—Coral Micro, Raspberry Pi, and Jetson Orin-and are summarized in Table \ref{tab:all_results}. TOn the Coral Micro, the results show that for CIFAR-10, AlexNet underperforms compared to ResNet models across all performance metrics. For ImageNet-1K, the trained AlexNet model could not be loaded onto the device due to its size, even after modifying the board's memory regions via the linker script to allocate additional space. In contrast, all models were successfully executed on the Raspberry Pi. 
%Regarding Early Exit, the expected behaviour latency-wise was observed: the more samples exited early (via decreasing the confidence threshold), the lower the time per inference on-device. % The results for the Early Exit implementations without any offloading can be found in Figures \ref{fig:EE-HI}.
For Jetson Orin, an important observation is that operating at $15$W results in lower energy consumption compared to $7$W. This is because higher power enables lower latency, leading to greater overall efficiency. This phenomenon is similar to the trade-offs observed in BLE vs. Wi-Fi offloading times. \color{black}

For CIFAR-10, Jetson Orin is the most efficient device in terms of both energy consumption and latency compared to the other two devices. While the Raspberry Pi is less power-efficient than the Coral Micro for ResNet-8 ($\times$2.9) and ResNet-56 ($\times$5.6), an anomaly is observed with AlexNet, which is significantly more efficient on the Raspberry Pi ($\times$5.8) than on the Coral Micro.

For ImageNet-1K, a similar pattern emerges. However, when considering latency, the Raspberry Pi outperforms the Coral Micro for ResNet-50 by a factor of 1.5. This suggests that running larger models on the Coral Micro incurs additional processing time, even with the assistance of its TPU accelerator.

\section{On-Device vs. HI vs. Remote Inference}
\label{sec:comparison}

\begin{figure}[ht!]
\centering
\begin{subfigure}{0.49\columnwidth}
    \includegraphics[width=\textwidth]{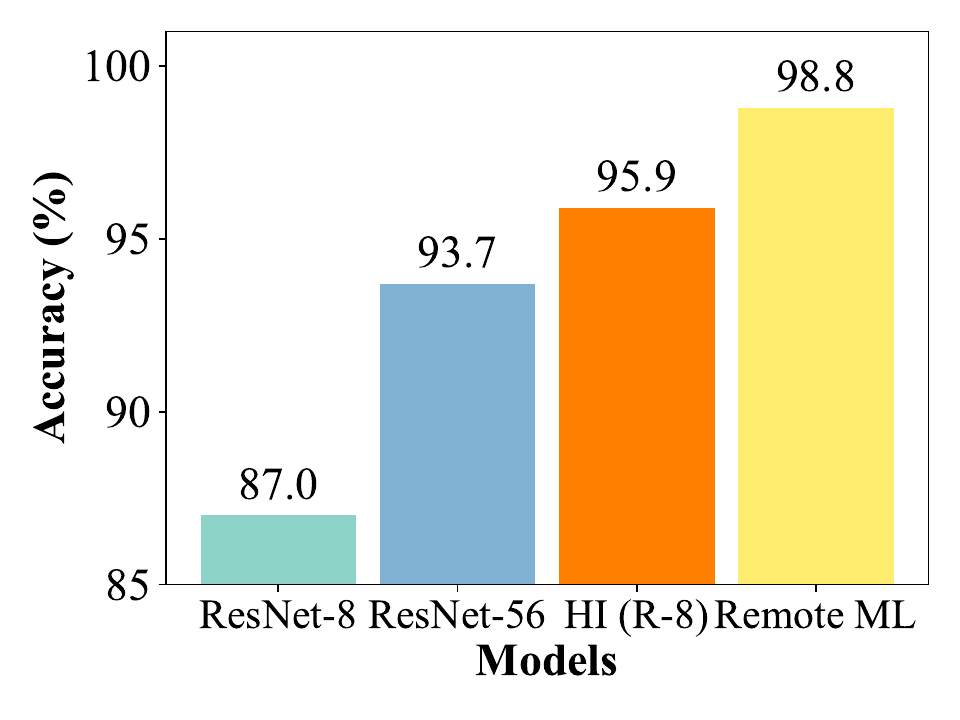}
    \caption{Accuracy on CIFAR-10}
    \label{fig:acc_C10}
\end{subfigure}
\hfill
\begin{subfigure}{0.49\columnwidth}
    \includegraphics[width=\textwidth]{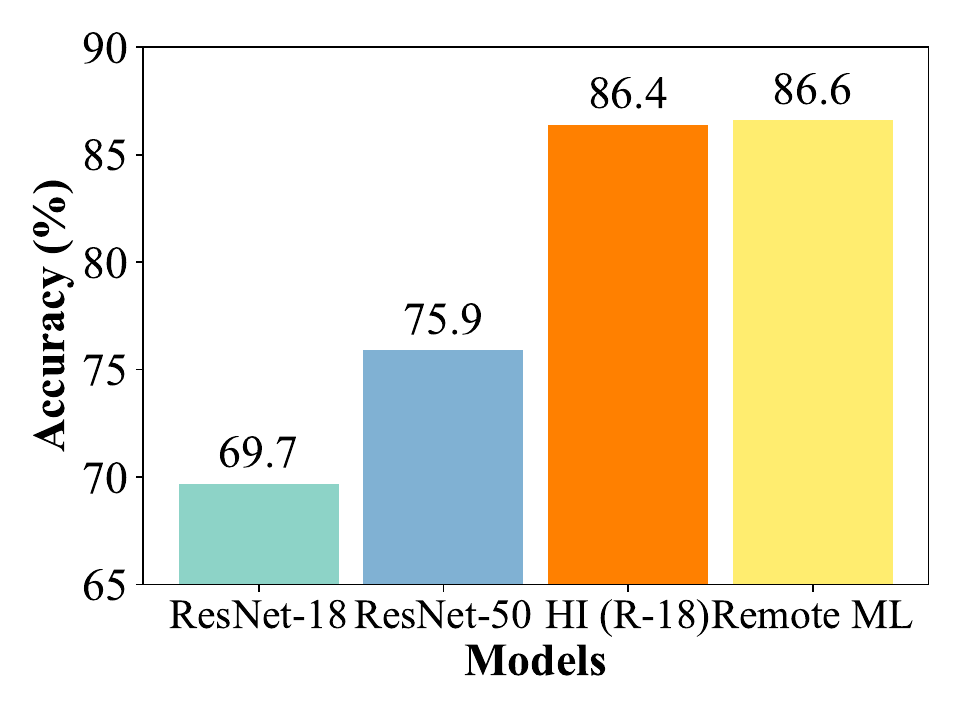}
    \caption{Accuracy on ImageNet-1K} 
    \label{fig:acc_IN}
\end{subfigure} 
\caption{Accuracy of different strategies on CIFAR-10 and ImageNet-1K datasets }
\label{fig:acc}
\end{figure}

\begin{figure*}
\centering
\begin{subfigure}[b]{0.48\textwidth}
    \includegraphics[width=\textwidth]{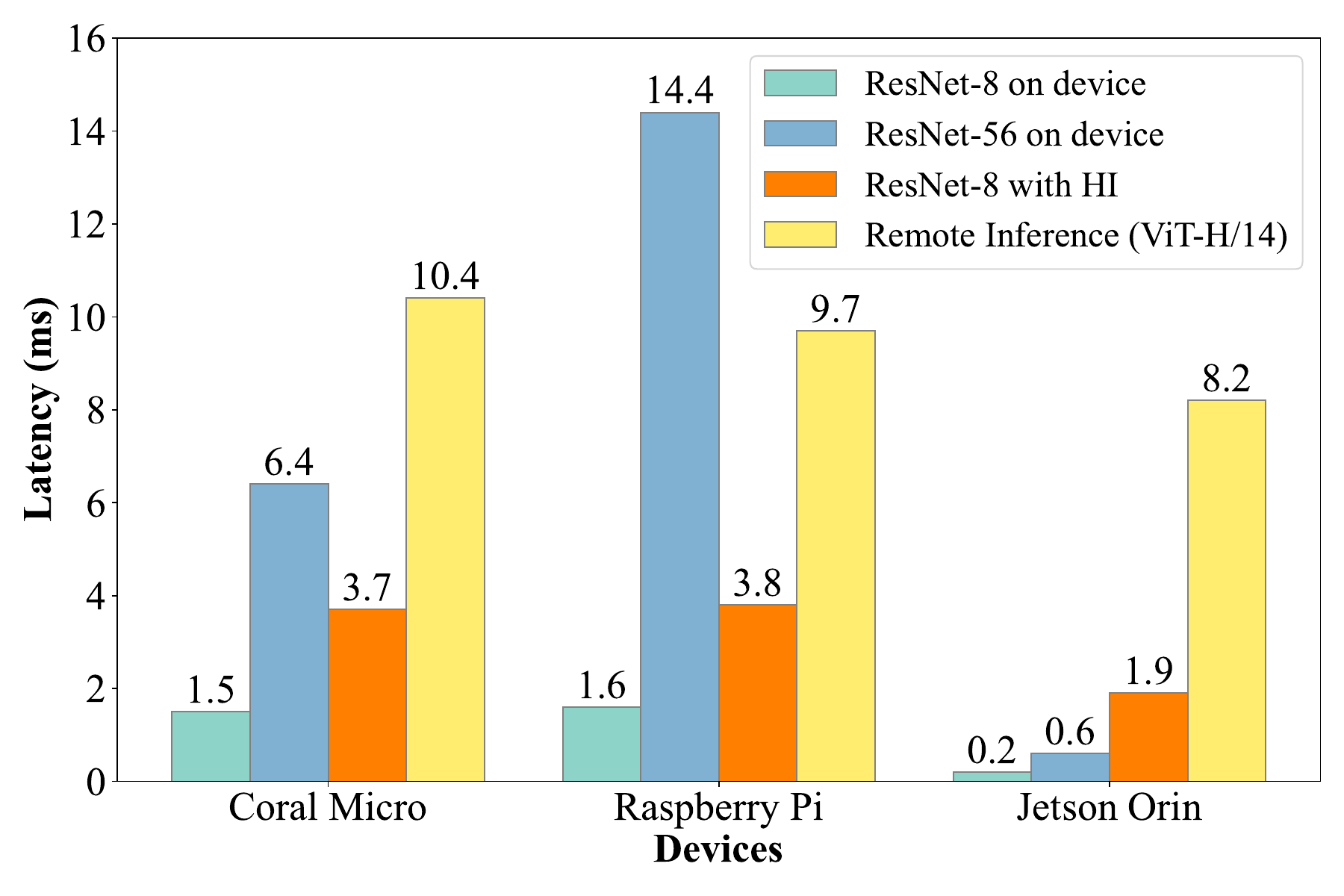}
    \caption{Latency for CIFAR-10}
    \label{fig:lat_C10}
\end{subfigure}
\hfill
\begin{subfigure}[b]{0.48\textwidth}
    \includegraphics[width=\textwidth]{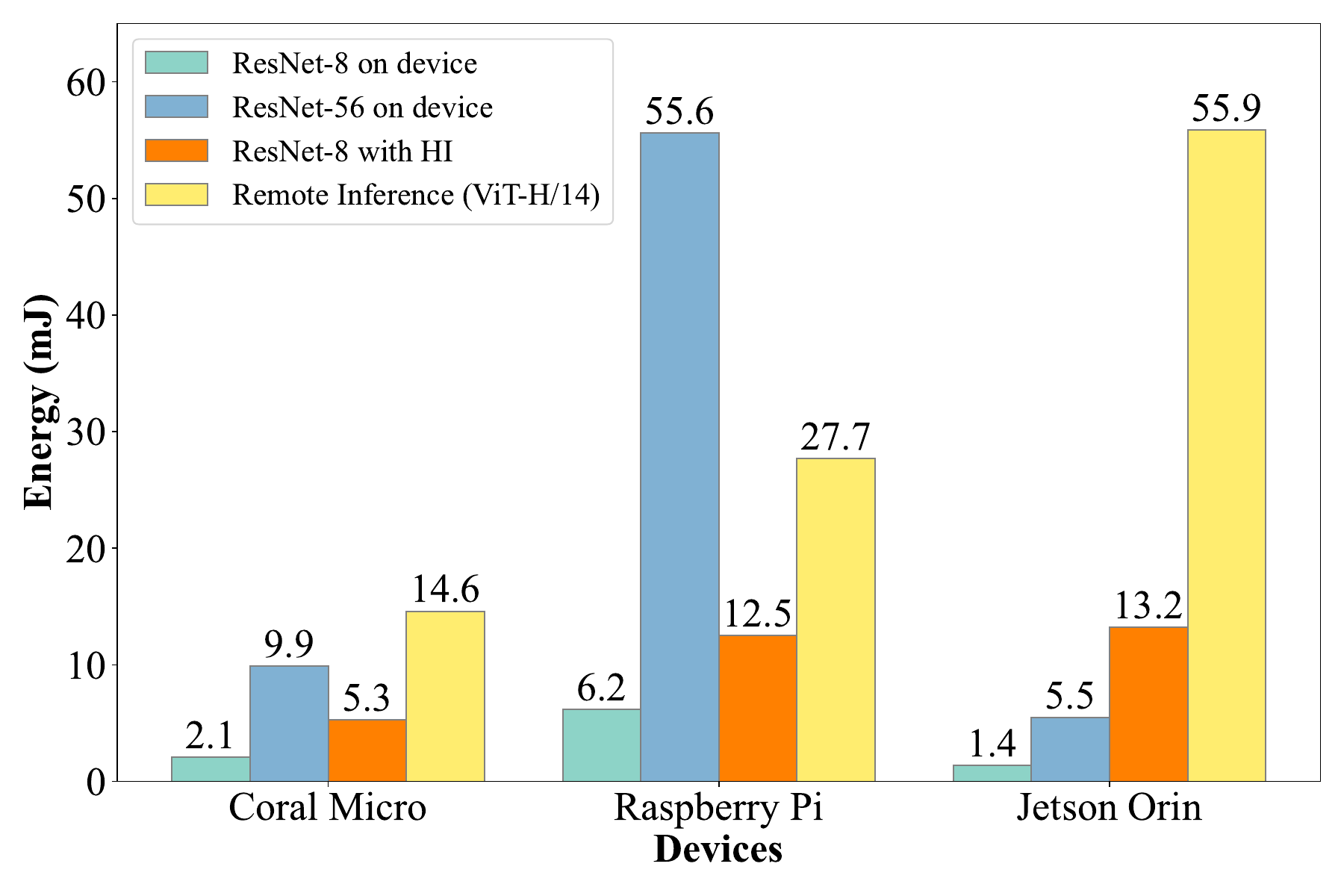}
    \caption{Energy consumption for CIFAR-10} 
    \label{fig:ene_C10}
\end{subfigure} 
\caption{Latency and energy comparison of on-device and HI systems for CIFAR-10 on different devices.}
\label{fig:C10}
\end{figure*}

\begin{figure*}
\centering

\begin{subfigure}{0.48\textwidth}
    \includegraphics[width=\textwidth]{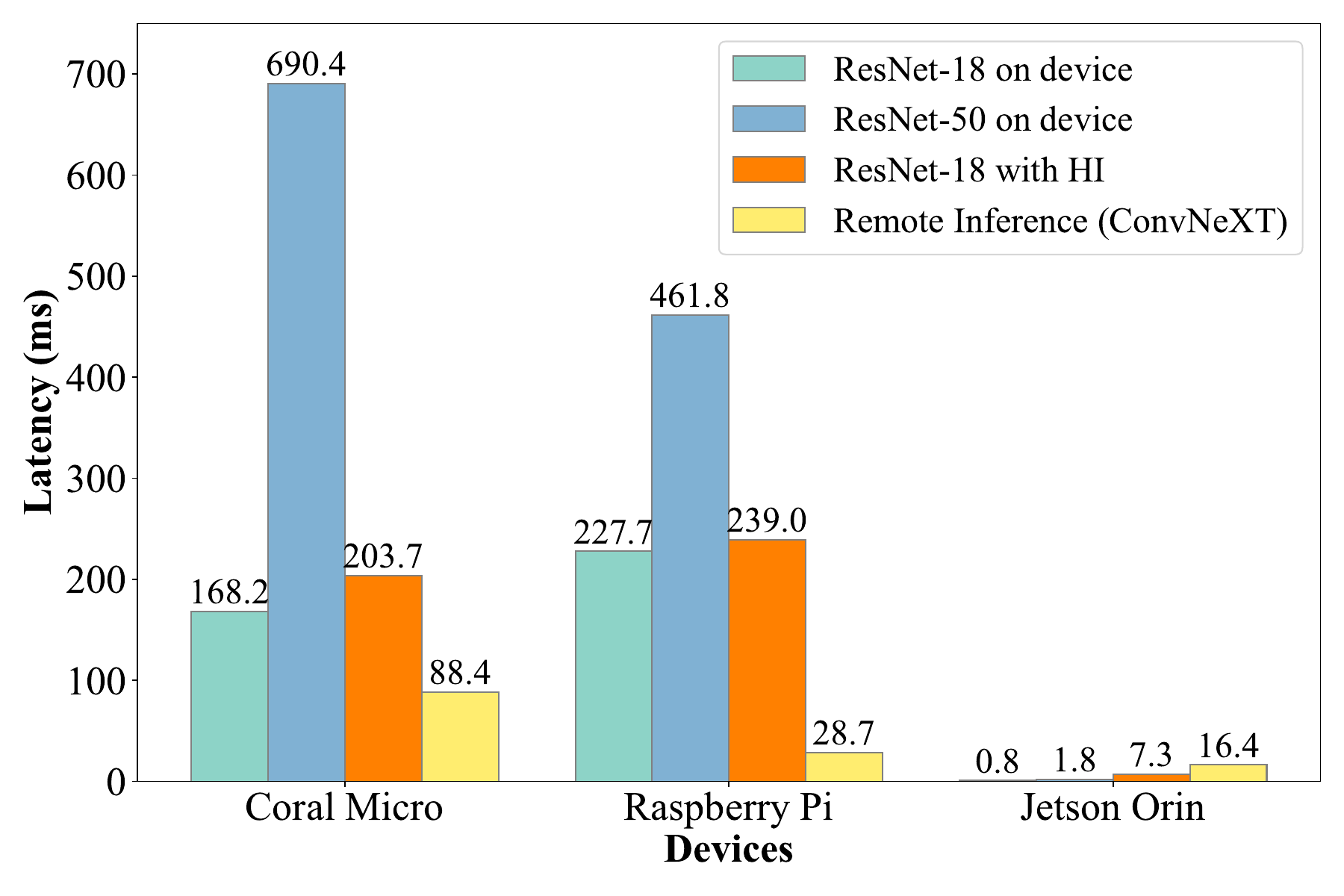}
    \caption{Latency for ImageNet-1K}
    \label{fig:lat_IN}
\end{subfigure}
\hfill
\begin{subfigure}{0.48\textwidth}
    \includegraphics[width=\textwidth]{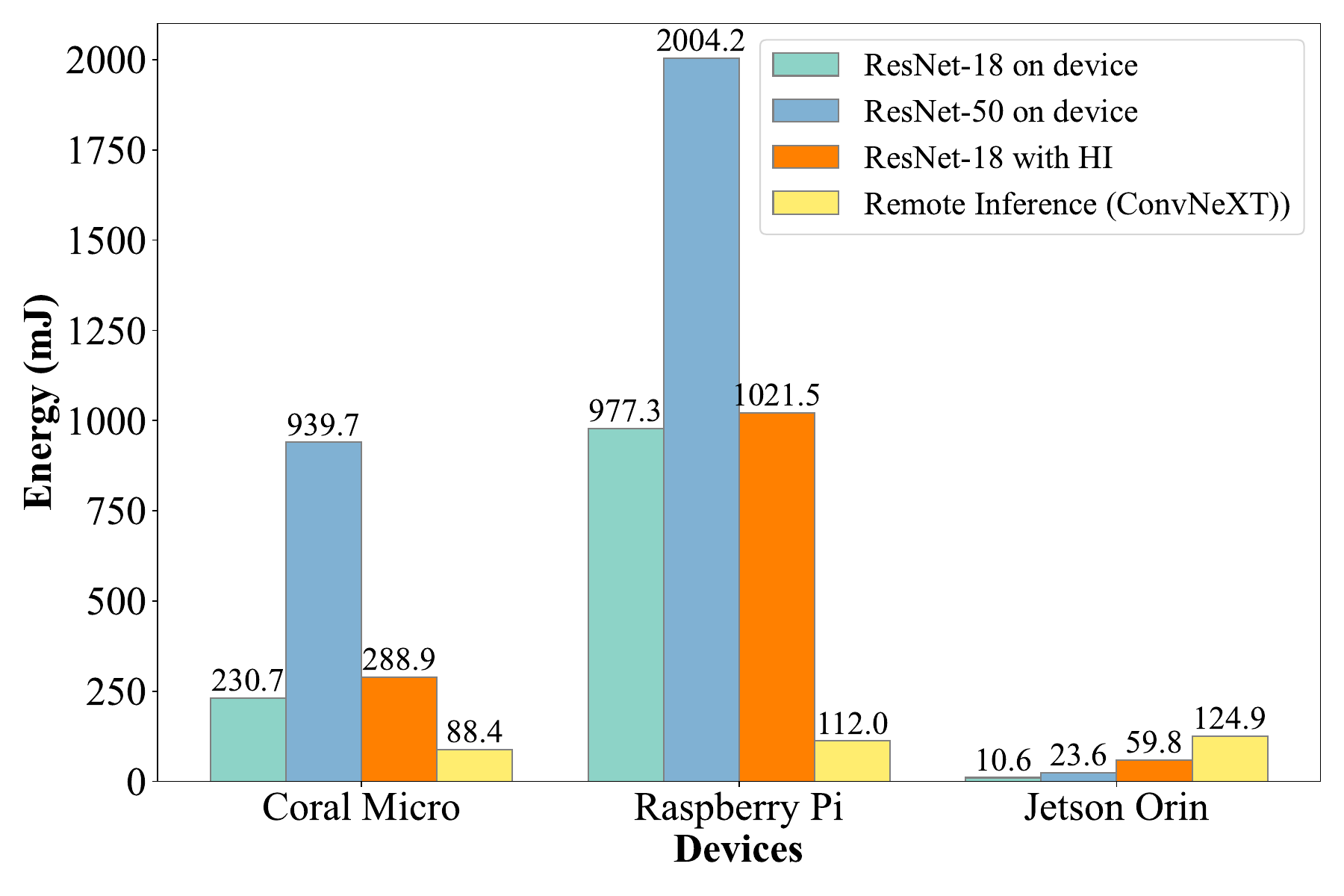}
    \caption{Energy consumption for ImageNet-1K} 
    \label{fig:ene_IN}
\end{subfigure} 
\caption{Latency and energy comparison of on-device and HI systems for ImageNet-1K on different devices.}
\label{fig:IN}
\end{figure*}

In Fig. \ref{fig:acc}, we present the accuracy achieved by different strategies for CIFAR-10 and ImageNet-1K datasets, respectively. The corresponding latency and energy consumption on on Coral Micro, Raspberry Pi, and Jetson Orin in Figs. \ref{fig:C10} and Fig. \ref{fig:IN}, respectively. ResNet-8 and ResNet-18, the state-of-the-art (SOTA) tinyML models for CIFAR-10 and ImageNet-1K, respectively, are used as the local DL models in the HI system. From Figs. \ref{fig:C10} and \ref{fig:IN}, we observe similar trends across both datasets for Coral Micro and Raspberry Pi. In all cases, the smallest local models—ResNet-8 for CIFAR-10 and ResNet-18 for ImageNet-1K—consume the least energy and exhibit the lowest latency. However, these models also deliver the lowest accuracies (cf. Fig. \ref{fig:acc}) and fail to meet the accuracy QoS requirements. 
%Larger local ML models improves accuracy but incurs substantially higher latency and energy consumption. 

%In Fig. \ref{fig:acc}, we present the accuracy, latency, and energy achieved by different strategies on Coral Micro, Raspberry Pi, and Jetson Orin for CIFAR-10 and ImageNet-1K datasets. We use ResNet-8, ResNet-18,  as the local DL models for the HI system for CIFAR-10 and ImageNet-1K, respectively. From Figs. \ref{fig:PC_coralc10}, \ref{fig:PC_coralIN}, \ref{fig:PC_rpic10}, and \ref{fig:PC_rpiIN}, we observe similar trends across the CIFAR-10 and ImageNet-1K datasets for both Coral Micro and Raspberry Pi. In all the subfigures, the smallest local ML models, ResNet-8 for CIFAR-10 and ResNet-18 for ImageNet-1K, consume the least energy and have the lowest latency among the models and the lowest accuracy. A larger local ML model improves accuracy but incurs substantially higher latency and energy consumption. 

The HI system achieves higher accuracy—up to $2.2$\% higher for CIFAR-10 (cf. Fig. \ref{fig:acc_C10}) and $11.5$\% for ImageNet-1K (cf. Fig. \ref{fig:acc_IN})—compared to both on-device models. This improvement comes with only a slight increase in latency and energy consumption relative to the smallest model (cf. Figs. \ref{fig:C10} and \ref{fig:IN}). 
%and significantly lower latency and energy consumption (up to 77.42\% less for CIFAR-10 and up to 70.5\% less for ImageNet-1K) compared to the larger on-device model.
Remote ML inference yields the highest accuracy for CIFAR-10 but results in significantly higher latency and energy consumption on both Coral Micro and Raspberry Pi, as shown in Figs. \ref{fig:lat_C10} and \ref{fig:ene_C10}. For ImageNet-1K, however, offloading data for remote ML inference leads to lower energy consumption and latency on non-GPU devices, due to the increased size of local ML models and the complexity of the dataset (cf. Figs. \ref{fig:lat_IN} and \ref{fig:ene_IN}). In contrast, Jetson Orin, with its GPU acceleration, is well-suited for on-device inference, as larger models like ResNet-56 for CIFAR-10 and ResNet-50 for ImageNet demonstrate significantly lower latency and energy consumption compared to remote ML inference (cf. Figs. \ref{fig:C10}, and \ref{fig:IN}).

Next, we compare the performance of three systems—on-device inference, the HI system, and remote inference—based on the QoS requirements defined in Section \ref{subsec:QoS}. For Arduino Nano and ESP32, in Fig. \ref{fig:PC_MCUs} shows that on-device inference using LR meets the accuracy QoS for the MNIST dataset, with significantly lower latencies compared to offloading. for CIFAR-10, the state-of-the-art tinyML model ResNet-8 fails to meet the $90$\% QoS requirement, necessitating the use of a larger model. The high latencies observed for ResNet-8 on these MCUs indicate that larger models would result in even greater latency and energy consumption, making remote inference the preferred strategy for meeting QoS requirements.

%Extremely resource-constrained MCUs like Arduino Nano and ESP32 perform well on the MNIST dataset, achieving the QoS requirement of 89.8\% with a simple LR model at 91.37\% accuracy and very low latencies of 0.87ms and 0.59ms, respectively. Offloading time and energy for these devices are significantly higher than on-device times (around 100ms), making on-device inference the preferred strategy for the MNIST dataset. For the CIFAR-10 dataset, the SOTA tinyML model ResNet-8 achieves 87\% accuracy, which falls below the accuracy QoS requirement $88.9$\%, necessitating a larger on-device model, but the high latencies for ResNet-8 on both MCUs (cf. Table \ref{mcus_c10}) imply that larger models would have even higher latency and energy requirements; thus, remote inference should be the preferred strategy to meet QoS requirements for these devices.

%For moderately powerful devices such as the Coral Micro, Raspberry Pi, and Jetson Orin, latency and energy for on-device inference are significantly lower, as can be observed in Fig. \ref{PC_total}. 
For Coral Micro and Raspberry Pi, we observe from Figs. \ref{fig:lat_C10} and \ref{fig:ene_C10} that for the CIFAR-10 dataset, both the on-device ResNet-56 model and the HI system with ResNet-8 can achieve the required accuracy QoS of $90$\%. Although the HI system introduces some latency and energy overheads compared to the on-device ResNet-8 model, these are minimal when compared to the significantly higher latency and energy consumption required for on-device inference with ResNet-56. As a result, the HI system meets the accuracy QoS with up to $73$\% lower latency and $77$\% less energy than the on-device model. Given the latency and energy QoS requirements (cf. Table \ref{off_table}), the HI system also achieves an absolute accuracy gain of over $2$\%. For Jetson Orin, on-device inference with ResNet-56 is much faster and more energy-efficient than the HI system. Therefore, for the $90$\% accuracy requirement, on-device inference is sufficient. However, since the HI system achieves higher accuracy ($1.64$\% more) while still meeting the energy and time QoS requirements, it remains the best strategy in those scenarios.

\begin{table*}[ht!]
\centering
\caption{Preferred strategy for QoS requirements. 
%The entries in bold suggest none of the strategies could fulfill that particular QoS requirement, and the next best possible solution has been identified.
}
\label{Req_QoS}
\begin{tabular}{ |c|c|c|c|c| }
\hline
 \multicolumn{2}{|c|}{} & \multicolumn{3}{|c|}{Best strategy given QoS Requirement}\\
 \hline
 
\hline
		{\textbf{\begin{tabular}[c]{@{}c@{}}Dataset\end{tabular}}} &{\textbf{\begin{tabular}[c]{@{}c@{}}Device\end{tabular}}} & {\textbf{\begin{tabular}[c]{@{}c@{}}Accuracy $ \geq$ SOTA -$10$\%\end{tabular}}} & {\textbf{\begin{tabular}[c]{@{}c@{}}Latency $\leq 0.5\times$ offload\end{tabular}}}
		& {\textbf{\begin{tabular}[c]{@{}c@{}}Energy $\leq 0.5\times$ offload\end{tabular}}}  \\    &   & &&   \\ \hline 
 
%MNIST & Arduino Nano & On Device (LR) & On Device (LR) & On Device (LR) \\
% \hline

%MNIST & ESP 32 & On Device (LR) & On Device (LR) & On Device (LR) \\
% \hline

CIFAR-10 & Coral Micro & ResNet-8 + HI & ResNet-8 + HI & ResNet-8 + HI  \\
 \hline

CIFAR-10 & Raspberry Pi & ResNet-8 + HI & ResNet-8 + HI & ResNet-8 + HI  \\
 \hline
 
CIFAR-10 & Jetson Orin & On Device (ResNet-56) & ResNet-8 + HI & ResNet-8 + HI  \\
 \hline

ImageNet-1K & Coral Micro & Remote Inference & Remote Inference & Remote Inference  \\
 \hline

ImageNet-1K & Raspberry Pi & Remote Inference & Remote Inference & Remote Inference  \\
 \hline

ImageNet-1K & Jetson Orin & ResNet-18 + HI & ResNet-18 + HI & ResNet-18 + HI  \\
 \hline

\end{tabular}
\end{table*}

\begin{figure*}[ht!]
    \centering
    \includegraphics[scale=0.39]{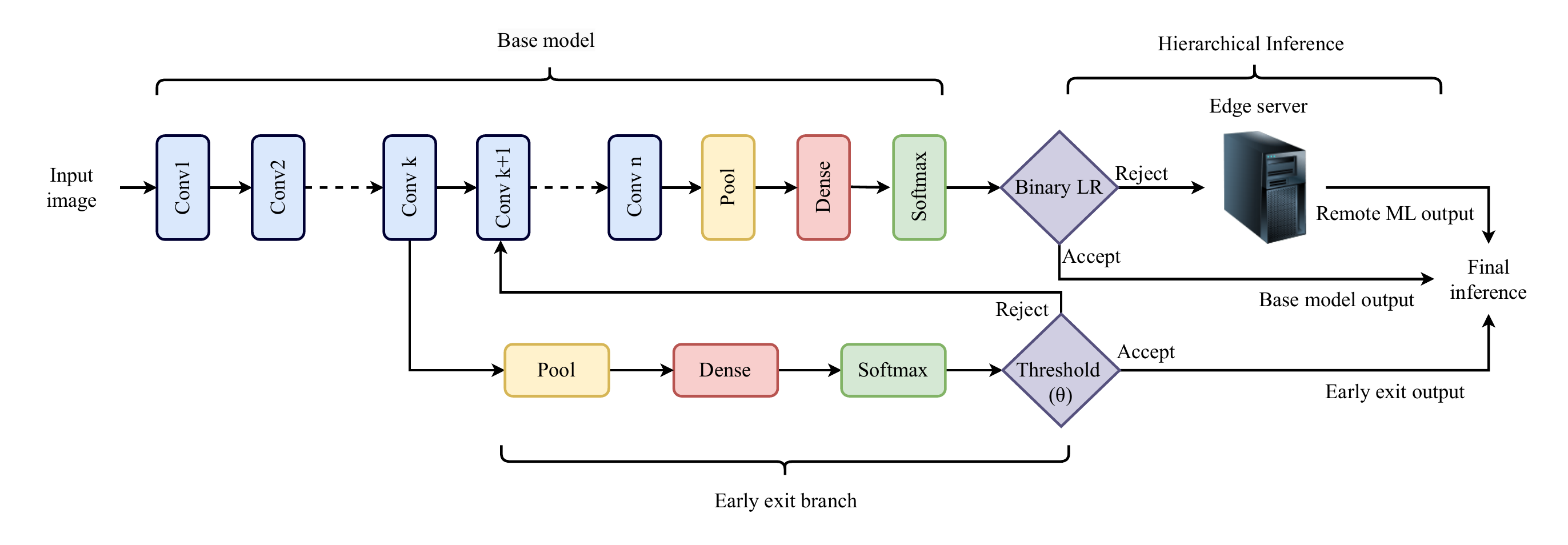}
    \caption{EE-HI for a Convolutional Neural Network (CNN) with a single EE branch.}
    \label{fig:EE-HI_arch}
\end{figure*}

From Table \ref{tab:all_results} and Fig. \ref{fig:acc_IN}, we observe that for the ImageNet-1K dataset, the on-device DL models ResNet-18 and ResNet-50, with accuracies of $69.7$\% and $76$\% respectively, fall well short of the required accuracy QoS of $80$\%. Furthermore, both models consume significantly more energy and exhibit higher latency on Coral Micro and Raspberry Pi compared to remote inference (cf. Figs. \ref{fig:lat_IN} and \ref{fig:ene_IN}). 
%Notably, for the larger on-device model ResNet-50 which achieves around $76$\% accuracy (lesser than the accuracy QoS requirement), latency and energy requirements on Coral Micro and Raspberry Pi are even higher.
This makes using a larger local model on these two devices infeasible. In contrast, offloading data samples for remote inference achieves higher accuracy with significantly lower latency and energy consumption. Therefore, remote inference is the optimal strategy for meeting the $80$\% accuracy QoS requirement. Additionally, it is the fastest and most energy-efficient solution for non-GPU devices.

%These findings are detailed in Table \ref{Req_QoS}. 
On Jetson Orin, however, the on-device inference is faster and more energy-efficient. Therefore, we choose the RegNetY32GF model for on-device inference, as both ResNet-18 and ResNet-50 fail to meet the $80$\% accuracy requirement. Using RegNetY32GF, we achieved $82.09$\% accuracy with an inference time of $14.88$ms and energy consumption of $213.67$mJ, which is higher than that of remote inference. In contrast, the HI system with ResNet-18 meets the accuracy QoS with $50$\% lower latency and $72$\% lesser energy. Further, HI improves accuracy by $2.14$\% over RegNetY32GF while satisfying both latency and energy QoS.
The best strategies for each QoS requirement are summarized in Table \ref{Req_QoS}.

It is noteworthy that the decision module plays a crucial role in the success of HI systems, as demonstrated in \eqref{acuracy_alt_eq}. In our experiments, the binary LR performed well, achieving F1 scores of 0.86 on ResNet-8 for CIFAR-10 and 0.83 on ResNet-18 for the ImageNet-1K dataset. A key insight from this performance comparison is that: 
%It is able to differentiate simple samples and complex samples with high accuracy minimizing false positives and false negatives.
%\vspace{-1mm}
\begin{itemize}
    \item In general, on resource-constrained devices, relatively complex tasks—such as CIFAR-10 on the Arduino Nano and ESP32, or ImageNet-1K on the Coral Micro and Raspberry Pi—demand significantly higher energy and latency to meet QoS requirements with on-device inference. As a result, offloading these tasks for remote inference proves to be a more efficient choice. %\textcolor{blue}{Can go after the first QoS paragraph.}
    \item If on-device inference is feasible, it is often more advantageous to choose an HI system with a smaller model that does not meet the QoS requirement rather than a larger on-device model that does. HI systems typically satisfy QoS requirements while optimizing latency and energy consumption, balancing trade-offs between accuracy and resource efficiency. %\textcolor{blue}{This needs to be at the end of the section}
\end{itemize}

{\allowdisplaybreaks
\section{Early Exit with HI}\label{sec:EEHI}

\subsection{Algorithm Description}
\begin{comment}
Recall that HI requires all the data samples to be first inferred using the local ML, which adds a fixed overhead to the latency and the energy (cf. \eqref{latency_eq} and \eqref{energy_eq}). One can reduce this overhead using the early exit technique \cite{teerapittayanon2016branchynet} for the local ML. To this end, in this section, we propose a novel strategy by combining early exit with the HI system, referred to by EE-HI. Using CIFAR-10 image classification on Raspberry PI and Jetson Orin, we present a case study on how EE-HI further lowers the latency and energy of HI given the $90$\% SOTA accuracy requirement. 
\end{comment}
From \eqref{latency_eq}, it can be inferred that the latency of an HI system depends on four factors: \emph{(i)} \textit{on-device model latency}, \emph{(ii)} \textit{HI decision module latency (binay LR in this case)}, \emph{(iii)} \textit{fraction of offloaded samples}, and \emph{(iv)} \textit{offloaded sample latency}, which accounts for both the communication time and the computation time at the server. Thus, reducing any of these factors will reduce the overall latency per inference in the HI system. The same conclusion applies to the energy consumption of the HI system (cf. \eqref{energy_eq}).
%According to inference strategies discussed so far, each sample must pass through the entire local ML model before being either accepted or offloaded. 
Recall that HI incurs latency and energy overhead due to the on-device model as every sample is first processed on it. To mitigate this overhead, we use the early exit technique \cite{teerapittayanon2016branchynet}, which has been proven to reduce on-device overhead while maintaining comparable accuracy. Specifically, we refer to the HI system using an on-device model (base model) with early exits as EE-HI system. An overview of an EE-HI system using CNNs is shown in Fig. \ref{fig:EE-HI_arch}.
%As existing techniques such as early exit \cite{teerapittayanon2016branchynet} have been shown to reduce on-device overhead while maintaining similar accuracy, we evaluate how introducing this strategy can enhance an HI system. As mentioned earlier, we refer to the case of early exit with HI as EE-HI. 

We design and train the base models with early exit branches, referred to as EE models, following the pioneering BranchyNet approach \cite{teerapittayanon2016branchynet}. 
Developing an EE model is a two-step process. First, an EE model with the desired architecture is created by adding branches with exit points between the base model's layers. An EE model with a single exit branch at layer $k$ is shown in Fig. \ref{fig:EE-HI_arch}. Note that an EE model possesses one input and multiple outputs—one for each exit branch introduced during the design phase and the final layer. Then, the EE model is trained with a joint loss function, calculated as a weighted sum of the individual loss functions for each exit branch. Let $C$ denote the set of output labels, and $w_n$ denote the relative weight assigned to the exit branch indexed by $n$. We determine the weights $w_n$ following the guidelines provided in the BranchyNet approach \cite{teerapittayanon2016branchynet}. Let $\textbf{y} = \{y_c: c\in C\}$ denote the ground truth vector, where $y_c \in \{0,1\}$ is $1$ for the ground truth label and $0$, otherwise. The output of the EE model is denoted by $\hat{\textbf{y}}_c$, a concatenation of the vectors $\hat{\textbf{y}}_{\text{exit}_n} = \{\hat{y}_{nc}: c \in C\}$, where $\hat{y}_{nc}$ is the predicted (confidence) value for class $c$ at exit $n$. Each branch's loss is defined as the \textit{Cross-Entropy} loss between $\textbf{y}$ and $\hat{\textbf{y}}_{\text{exit}_n}$. Thus, the joint loss function, denoted by $L_{\text{branchynet}}(\hat{\textbf{y}}, \textbf{y})$, that we use to train the EE model, is given by 
\begin{equation*}
    L_{\text{branchynet}}(\hat{\textbf{y}}, \textbf{y}) = \sum_{n=1}^{N} w_n L(\hat{\textbf{y}}_{\text{exit}_n}, \textbf{y}),
\label{total_loss}
\end{equation*}
where
\begin{equation*}
    L(\hat{\textbf{y}}_{\text{exit}_n}, \textbf{y}) = -\frac{1}{|C|} \sum_{c \in C} y_c \log \hat{y}_{nc}.
\label{branch_loss}
\end{equation*}
%the second step involves adding conditional statements for each branch, such that the EE model only has one output. 
% If the confidence output in a given branch for the predicted class is above the threshold value, this prediction is accepted, and inference is completed for the given sample. Otherwise, the inference would continue until the next branch. If no early branch prediction is accepted, the model will accept the backbone prediction..

Several variables must be considered when designing an EE architecture, as they significantly impact model performance. The most critical factors are the number of branches ($N$), their placement, and their composition.
%Along the line of past works \cite{laskaridis2021adaptive}\cite{teerapittayanon2016branchynet}, 
The placement and composition of the EE branches are determined through trial-and-error, following the guidelines established in previous works \cite{teerapittayanon2016branchynet,laskaridis2021adaptive}. Additionally, for each branch, it is necessary to define the confidence threshold that determines whether to accept the inference at the given exit or continue processing in subsequent layers. We identify the optimal threshold using a brute-force approach. The optimal parameter values for each model are shown in Table \ref{tab:ee_hi_parameters}.

For ResNet-8, we place one EE branch after the second residual block (or fifth convolutional layer), and for AlexNet, one EE branch is added after the second convolutional layer. For ResNet-56, two EE branches are used: the first is placed after layer $2$, and the second after layer $19$. Larger models can accommodate more branches without introducing excessive overhead relative to the base model's latency. As noted earlier, the brute-force approach is used to compute the optimal threshold combination that satisfies the QoS requirements for each scenario.

\begin{table}[h]
\caption{Early Exit Parameters in EE-HI}
\footnotesize
    \centering
    \begin{tabular}{|c|c|c|}
        \hline
        \textbf{Model} & \textbf{EE Branch Placement} & \textbf{Threshold ($\theta$)} \\
        \hline
        ResNet-8  & Layer 5 & $\theta = 0.69$ \\
        ResNet-56 & 1st: Layer 2, 2nd: Layer 19 & $\theta_1 = 0.81, \theta_2 = 0.84$ \\
        AlexNet   & Layer 2 & $\theta = 0.75$ \\
        \hline
    \end{tabular}
    \label{tab:ee_hi_parameters}
\end{table}

During the inference phase, once the model is successfully trained, a sample is first evaluated at the initial EE branch. If the confidence level (maximum softmax value) for classification exceeds the threshold, the system accepts the output and proceeds to the next sample. Otherwise, the sample continues to subsequent layers for further processing. This ensures that additional processing overhead is incurred only when necessary.

\begin{table}[ht!]
\tiny
\centering
\caption{ Results for the EE-HI system with optimal threshold
subject to the QoS requirement.} 
\begin{tabular}{|c|c|c|c|c|c|c|l|}
\hline
\textbf{Model} & \textbf{Device} & \textbf{Pow.} & \textbf{\begin{tabular}[c]{@{}c@{}}EE-HI \\ Time \\ (ms)\end{tabular}} & \textbf{\begin{tabular}[c]{@{}c@{}}EE-HI \\ Energy \\ (mJ)\end{tabular}} & \textbf{\begin{tabular}[c]{@{}c@{}}HI \\ Time \\ (ms)\end{tabular}} & \textbf{\begin{tabular}[c]{@{}c@{}}HI\\ Energy \\ (mJ)\end{tabular}} & \begin{tabular}[c]{@{}l@{}}\textbf{EE-HI}\\ \textbf{Threshold}\end{tabular} \\ \hline
\multirow{3}{*}{ResNet-8} & RaspPi &  & 2.71 & 9.00 & 3.82 & 12.55 & \multirow{3}{*}{\begin{tabular}[c]{@{}l@{}}$\theta = 0.69$\end{tabular}} \\ \cline{2-7}
 & \multirow{2}{*}{Jetson} & 7W & 1.57 & 9.70 & 2.04 & 12.57 &  \\ \cline{3-7}
 &  & 15W & 1.47 & 10.26 & 1.87 & 13.05 &  \\ \hline
\multirow{3}{*}{ResNet-56} & RaspPi &  & 6.13 & 23.02 & 15.21 & 58.10 & \multirow{3}{*}{\begin{tabular}[c]{@{}l@{}}$\theta_1 = 0.81$\\ $\theta_2=0.84$\end{tabular}} \\ \cline{2-7}
 & \multirow{2}{*}{Jetson} & 7W & 0.84 & 5.52 & 1.67 & 10.95 &  \\ \cline{3-7}
 &  & 15W & 0.69 & 5.22 & 1.24 & 9.63 &  \\ \hline
\multirow{3}{*}{AlexNet} & RaspPi &  & 5.66 & 19.47 & 7.81 & 28.22 & \multirow{3}{*}{\begin{tabular}[c]{@{}l@{}}$\theta = 0.75$\end{tabular}} \\ \cline{2-7}
 & \multirow{2}{*}{Jetson} & 7W & 2.69 & 17.12 & 3.27 & 21.59 &  \\ \cline{3-7}
 &  & 15W & 2.51 & 17.98 & 2.90 & 21.53 &  \\ \hline
\end{tabular}
\label{EEHI}
\label{tab: EE_thresholds}
\end{table}

\begin{figure}[ht!] % [H] means "here", i.e., place the figure exactly here
  \centering
  \begin{subfigure}[b]{0.5\textwidth}
    \centering
    \includegraphics[width=\textwidth]{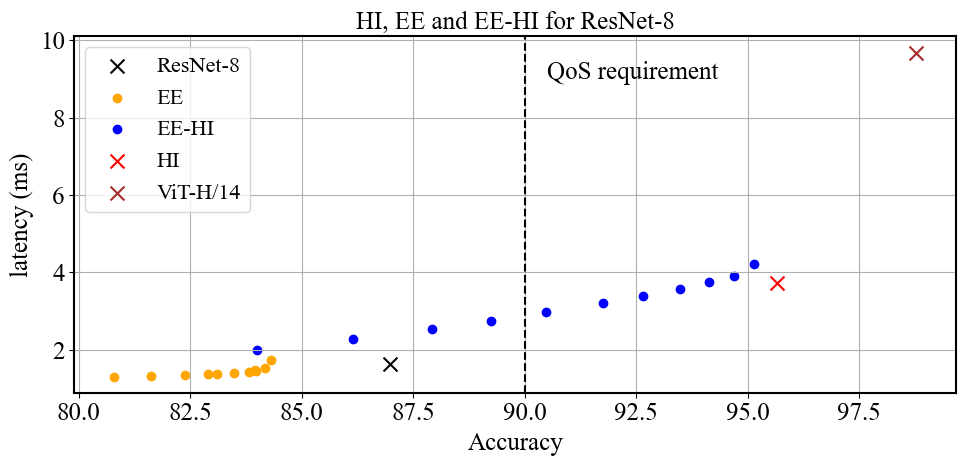}
    
    \label{fig:sub1}
  \end{subfigure}
  \hfill % to add some spacing between the subfigures
  \begin{subfigure}[b]{0.5\textwidth}
    \centering
    \includegraphics[width=\textwidth]{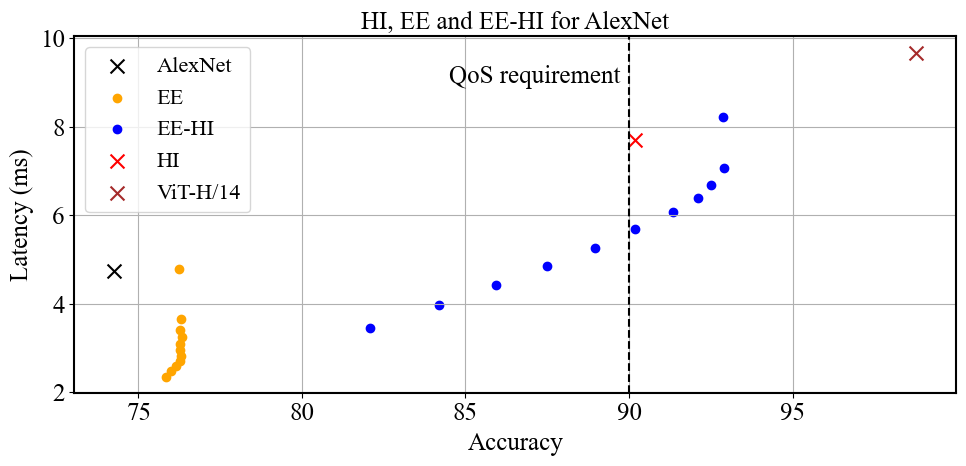}
    
    \label{fig:sub3}
  \end{subfigure}
  \hfill
  \begin{subfigure}[b]{0.5\textwidth}
    \centering
    \includegraphics[width=\textwidth]{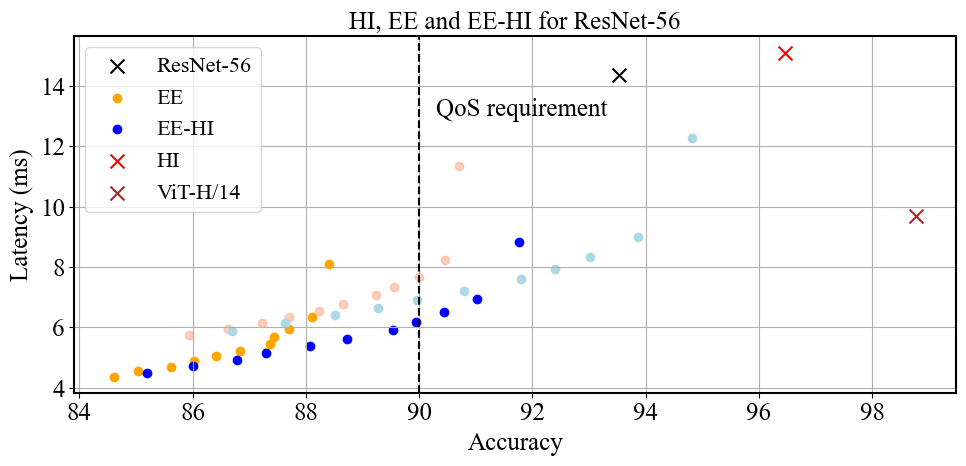}
    
    \label{fig:sub2}
  \end{subfigure}

  \caption{EE, HI, EE-HI comparison for different models for CIFAR-10 on Raspberry Pi.}

  \label{fig:EE-HI}
\end{figure}

\begin{comment}

\begin{figure}
    \centering
    \includegraphics[width=0.5\textwidth]
    {Inaki/EEresnet8.png}
    \caption{Comparison of different approaches with Resnet8 as base model for Raspberry}
    \label{fig:EEHI_Res8}
\end{figure}

\begin{figure}
    \centering
    \includegraphics[width=0.5\textwidth]
    {Inaki/EEalexnet.png}
    \caption{ Comparison of different approaches with
AlexNet as base model for Raspberry. A QoS of at least 90\% of the accuracy of the SOTA model is required.}
    \label{fig:EEHI_Alex}
\end{figure}

\begin{figure}
    \centering
    \includegraphics[width=0.5\textwidth]
    {Inaki/EEresnet562.png}
    \caption{ Comparison of different approaches with
ResNet56 as base model for Raspberry PI. A QoS of at least 90\% of the accuracy of the SOTA model is required. The dark blue and orange dots are obtained fixing $\theta_{1} = 0.77$ an varying $\theta_{2}$. The lighter colors are obtained fixing $\theta_{1} = 0.95$ and varying $\theta_{2}$, to show the flexibility of the model with multiple branches}
    \label{fig:EEHI_Res56}
\end{figure}

\end{comment}

\subsection{Results}
%In order to compare the performance of this novel strategy, a constraint needs to be set on either accuracy, latency, or energy consumption. This allows us to choose the optimal combination of early exit thresholds that satisfy the given conditions. 
We evaluated the HI and EE-HI systems on Raspberry Pi and Jetson Orin using the CIFAR-10 dataset. The minimum accuracy QoS requirement for CIFAR-10 is 90\%, and the results are summarized in Table \ref{EEHI}.
%We find the optimal threshold combination for each model using a brute-force approach. 
%The results are presented in Table \ref{EEHI}. 
For each model and device, EE-HI consistently outperforms the base model with the HI strategy, achieving reductions of up to $60$\% in both latency and energy consumption for ResNet-56 on Raspberry Pi.

A more in-depth comparison is presented in Fig. \ref{fig:EE-HI}, where the performance of all tested strategies is depicted: base model, base model with HI using binary LR (HI), EE (base model with early exit), EE-HI using binary LR (EE-HI), and remote inference with a larger DL model (ViT-H/14). For the EE and EE-HI strategies, we vary the confidence thresholds from $0.5$ to $1$, incrementally increasing the threshold. Higher thresholds result in fewer samples being accepted early, leading to improved accuracy but increased latency. This highlights the trade-off between latency/energy and accuracy. The optimal thresholds, where the dotted lines intersect the QoS requirement limit, are provided in Table \ref{EEHI} for EE-HI.

Note that EE models (orange dots) fail to meet the accuracy QoS requirement, regardless of the selected threshold. Additionally, they achieve lower accuracy than the base models (black cross marker), except for AlexNet. However, with appropriate threshold selection, EE models can provide significant improvements in latency and energy consumption while maintaining reasonable accuracy. Introducing more branches in the model increases flexibility in the accuracy-latency trade-off, as demonstrated by the ResNet-56 implementation in Fig. \ref{fig:EE-HI}. The dark-colored lines in the figure represent all possible configurations attainable when the first threshold is fixed at $\theta_{1} = 0.81$ (with $\theta_{2}$ increasing monotonically). To further illustrate the concept, two light-colored lines are included, showing configurations obtained by fixing $\theta_{1} = 0.95$ and varying $\theta_{2}$. While these configurations do not achieve the optimal QoS, they demonstrate the ability to achieve higher accuracy while still benefiting from latency reductions. As a result, EE-HI emerges as the best strategy across all the considered scenarios. It combines the accuracy improvements of HI with the reduced latency and energy benefits of EE models. Moreover, its flexibility enables the development of systems tailored to specific tasks and QoS requirements.

}

\section{Conclusion and Future Work}\label{sec:conclusion}
In this work, we presented processing times, energy consumption, and accuracy measurements for inference using AlexNet and ResNet variants on five devices: Arduino Nano, ESP32, Coral Micro, Raspberry Pi, and Jetson Orin, across MNIST, CIFAR-10, and ImageNet image classification tasks. Using these measurements, we compared the performance of on-device inference, remote inference, and HI approaches. Unlike existing works, we propose that a HI system with a smaller on-device model is more effective for meeting QoS requirements than a HI system using a SOTA on-device model. Our results show that the HI system complements on-device inference, achieving up to a $73$\% reduction in latency and a $77$\% reduction in on-device energy consumption in the studied scenarios. Furthermore, we designed a hybrid EE-HI system that integrates EE models into the HI framework. This approach demonstrated up to a $60$\% reduction in latency and energy consumption compared to HI alone for CIFAR-10 image classification on Raspberry Pi and Jetson Orin. %it further reduces the latency and energy consumption by more than $50$\% compared to HI for CIFAR-10 image classification on Raspberry Pi and Jetson Orin Nano. 
Despite these gains, we find that in scenarios where no on-device model is small enough to meet latency and energy QoS requirements while maintaining reasonable accuracy, the HI system defaults to offloading all data samples for remote inference.
%Despite these gains, using HI remains inferior to remote inference in certain scenarios where no on-device models are small enough to meet latency and energy QoS requirements while maintaining reasonable accuracy.
This underscores the need for designing DL models with a better accuracy-to-size ratio than current state-of-the-art (SOTA) on-device models.

Our work opens several new research directions. For the early exit technique, we employed the BranchyNet design approach \cite{teerapittayanon2016branchynet}. 
%Although pioneering, BranchyNet is not the most sophisticated approach. 
We aim to explore novel techniques like self-distillation \cite{zhang2021self}, which have shown significant improvements in early exit implementation by reducing latency and energy consumption while improving accuracy compared to the base model. We also aim to explore more efficient optimization techniques to streamline threshold selection in EE-HI.
%We also plan to investigate how this concept could identify complex samples earlier in the network, further accelerating system inference. 
%The HI decision module is crucial for achieving higher accuracy and lower latency and energy consumption. It determines the number of samples to be offloaded for further inference. 
While binary LR is an effective HI offloading decision algorithm, it produces false positives (incorrect local inference samples that are not offloaded) and false negatives (correct local inference samples that are offloaded). In our HI systems, binary LR achieved an F1 score of 0.86 for ResNet-8 on CIFAR-10 and 0.83 for ResNet-18 on ImageNet-1K.
%The F1 score measures how well the decision module minimizes false positives and false negatives, depending on both the decision module and the performance of the local ML model. 
To further enhance the efficiency of HI systems, we plan to investigate alternative offloading decision algorithms that reduce false positives and false negatives. Additionally, developing local ML models with a high accuracy-to-size ratio will enable these models to meet stringent QoS requirements with HI while maintaining low latency and energy consumption.
Finally, our experiments were conducted in a static setup with a dedicated server and stable WiFi access, resulting in consistent offloading times. Future work will focus on studying the impact of device mobility and network interference on offloading times and the overall efficiency of HI systems.
%introduces variability in communication latency. 
%This variability can lead to sub-optimal performance in HI systems, affecting both latency-sensitive applications and energy efficiency.

%\clearpage
%\input{To_Do.tex}
%\newpage
\section*{Appendix}

\section*{Additional measurements}
In Table \ref{mcus_c10}, we present the accuracy and latency measurements taken for Arduino Nano and ESP-32 for CIFAR-10 dataset. As mentioned before, these MCUs possess very limited computational capabilities making it hard to satisfy latency requirements for even moderately complex tasks. The average latency for a single on-device inference is approximately $6.5$ times slower than that of remote inference for the Arduino Nano, and nearly $60$ times slower for the ESP-32. It should be noted that remote inference is significantly faster on the ESP-32 because it uses Wi-Fi as its communication protocol, as opposed to Bluetooth used by the Arduino Nano. Therefore, the preferred solution would be to offload the inference task to a powerful edge server and receive the corresponding prediction. This offloading process is labeled as ``Remote" in Table \ref{mcus_c10}. The term ``HI" refers to the entire system, encompassing both on-device inference and remote inference for a specified subset of images, denoted as ``\% offload," which indicates the percentage of total images that are offloaded for more detailed inference.

\begin{table}[ht]
\centering
\caption{ResNet-8 on CIFAR-10 results for Arduino Nano and ESP-32. }
\begin{tabular}{ll|ll|}
\cline{3-4}
                                                         &                       & \multicolumn{1}{l|}{\textbf{Arduino}} & \textbf{ESP-32} \\ \hline
\multicolumn{1}{|l|}{\multirow{2}{*}{\textbf{Accuracy}}} & \textbf{Device}       & \multicolumn{2}{l|}{87.07}                              \\ \cline{2-4} 
\multicolumn{1}{|l|}{}                                   & \textbf{HI (\% offload)} & \multicolumn{2}{l|}{95.75 (21.57)}                      \\ \hline
\multicolumn{1}{|l|}{\multirow{2}{*}{\textbf{Latency (ms)}}}      & \textbf{Device}       & \multicolumn{1}{l|}{1230.48}          & 602.52          \\ \cline{2-4} 
\multicolumn{1}{|l|}{}                                   & \textbf{Remote} & \multicolumn{1}{l|}{190.35}                 &       10.10          \\ \hline
\end{tabular}
\label{mcus_c10}
\end{table}

In Table \ref{tab:my-table}, we present the latency and accuracy measurements for the base models on the Jetson Orin Nano with different precision levels. The entire test set for CIFAR-10 and the entire validation set for ImageNet-1k have been used for accuracy computation. For latency measurement, the values are averaged over $10000$ images, that are randomly selected. This table reinforces the effects of quantization from a precise measurement standpoint. We can observe that as models are further quantized, the average latency requirements for a single inference reduce gradually without significantly impacting accuracy. Another key observation is that quantization strategies are more beneficial for larger models compared to smaller ones. For example, with the CIFAR-10 dataset, full integer quantization results in a $19$\% reduction in latency for ResNet-8 compared to full precision. In contrast, ResNet-56 and AlexNet experience $42$\% and $53$\% reduction, respectively. For the ImageNet-1K dataset, however, all three models exhibit similar gains in latency reduction, each achieving around $69$\%.

\begin{table}[ht]
    \centering
    \caption{Latency and accuracy for the base models on the Jetson Orin Nano with different precision levels.}
\label{tab:my-table}
\resizebox{\columnwidth}{!}{%
\begin{tabular}{ccc|cc|cc|cc|}
\cline{4-9}
                                                         &                                                         &                          & \multicolumn{2}{c|}{\textbf{FP32}}       & \multicolumn{2}{c|}{\textbf{FP16}}       & \multicolumn{2}{c|}{\textbf{INT8}}       \\ \cline{3-9} 
                                                         & \multicolumn{1}{c|}{\textbf{}}                          & \textbf{Power Mode}      & \multicolumn{1}{c|}{15 W}     & 7 W      & \multicolumn{1}{c|}{15 W}     & 7 W      & \multicolumn{1}{c|}{15 W}     & 7 W      \\ \hline
\multicolumn{1}{|c|}{\multirow{6}{*}{\textbf{CIFAR}}}    & \multicolumn{1}{c|}{\multirow{2}{*}{\textbf{ResNet8}}}  & \textbf{Accuracy ($\%$)} & \multicolumn{2}{c|}{87.19}               & \multicolumn{2}{c|}{87.16}               & \multicolumn{2}{c|}{86.86}               \\ \cline{3-9} 
\multicolumn{1}{|c|}{}                                   & \multicolumn{1}{c|}{}                                   & \textbf{Latency (ms)}    & \multicolumn{1}{c|}{0.212785} & 0.306191 & \multicolumn{1}{c|}{0.183042} & 0.260383 & \multicolumn{1}{c|}{0.171250} & 0.249488 \\ \cline{2-9} 
\multicolumn{1}{|c|}{}                                   & \multicolumn{1}{c|}{\multirow{2}{*}{\textbf{ResNet56}}} & \textbf{Accuracy ($\%$)} & \multicolumn{2}{c|}{93.75}               & \multicolumn{2}{c|}{93.76}               & \multicolumn{2}{c|}{93.65}               \\ \cline{3-9} 
\multicolumn{1}{|c|}{}                                   & \multicolumn{1}{c|}{}                                   & \textbf{Latency (ms)}    & \multicolumn{1}{c|}{1.102070} & 1.665560 & \multicolumn{1}{c|}{0.762705} & 1.150450 & \multicolumn{1}{c|}{0.640057} & 1.035910 \\ \cline{2-9} 
\multicolumn{1}{|c|}{}                                   & \multicolumn{1}{c|}{\multirow{2}{*}{\textbf{AlexNet}}}  & \textbf{Accuracy ($\%$)} & \multicolumn{2}{c|}{74.52}               & \multicolumn{2}{c|}{74.52}               & \multicolumn{2}{c|}{74.63}               \\ \cline{3-9} 
\multicolumn{1}{|c|}{}                                   & \multicolumn{1}{c|}{}                                   & \textbf{Latency (ms)}    & \multicolumn{1}{c|}{0.844494} & 1.759410 & \multicolumn{1}{c|}{0.551228} & 0.959152 & \multicolumn{1}{c|}{0.389317} & 0.627006 \\ \hline
\multicolumn{1}{|c|}{\multirow{6}{*}{\textbf{ImageNet}}} & \multicolumn{1}{c|}{\multirow{2}{*}{\textbf{ResNet18}}} & \textbf{Accuracy ($\%$)} & \multicolumn{2}{c|}{70}                  & \multicolumn{2}{c|}{69.812}              & \multicolumn{2}{c|}{69.592}              \\ \cline{3-9} 
\multicolumn{1}{|c|}{}                                   & \multicolumn{1}{c|}{}                                   & \textbf{Latency (ms)}    & \multicolumn{1}{c|}{2.74285}  & 6.81942  & \multicolumn{1}{c|}{1.26768}  & 2.81547  & \multicolumn{1}{c|}{0.8375}   & 1.85074  \\ \cline{2-9} 
\multicolumn{1}{|c|}{}                                   & \multicolumn{1}{c|}{\multirow{2}{*}{\textbf{ResNet50}}} & \textbf{Accuracy ($\%$)} & \multicolumn{2}{c|}{76.088}              & \multicolumn{2}{c|}{76.078}              & \multicolumn{2}{c|}{76.010}              \\ \cline{3-9} 
\multicolumn{1}{|c|}{}                                   & \multicolumn{1}{c|}{}                                   & \textbf{Latency (ms)}    & \multicolumn{1}{c|}{5.82106}  & 13.5075  & \multicolumn{1}{c|}{2.8421}   & 6.32683  & \multicolumn{1}{c|}{1.80605}  & 4.03902  \\ \cline{2-9} 
\multicolumn{1}{|c|}{}                                   & \multicolumn{1}{c|}{\multirow{2}{*}{\textbf{AlexNet}}}  & \textbf{Accuracy ($\%$)} & \multicolumn{2}{c|}{56.544}              & \multicolumn{2}{c|}{56.534}              & \multicolumn{2}{c|}{56.452}              \\ \cline{3-9} 
\multicolumn{1}{|c|}{}                                   & \multicolumn{1}{c|}{}                                   & \textbf{Latency (ms)}    & \multicolumn{1}{c|}{4.58609}  & 8.41834  & \multicolumn{1}{c|}{2.38455}  & 4.23303  & \multicolumn{1}{c|}{1.47918}  & 2.63732  \\ \hline
\end{tabular}
    }
\end{table}

\section*{Acknowledgement}
This work is supported by 
\begin{itemize}
    \item The European Union through Marie Skłodowska-Curie Actions - Postdoctoral Fellowships (MSCA-PF) Project ‘‘DIME: Distributed Inference for Energy-efficient Monitoring at the Network Edge’’ under Grant 101062011 and by the "European Union – Next Generation EU".
    \item The Ministry of Employment and Social Economy of Spain through the Spanish Recovery, Transformation and Resilience Plan, in the framework of “Programa Investigo” call.
    \item MAP-6G (TSI-063000-2021-63) and RISC-6G (TSI-063000-2021-59) granted by the Ministry of Economic Affairs and Digital Transformation and the European Union-Next Generation EU through the UNICO-5G R\&D program of the Spanish Recovery, Transformation and Resilience Plan.
    \item Business Finland project Digital Twinning of Personal Area Networks for Optimized Sensing and Communication (8782/31/2022)
\end{itemize}

\bibliographystyle{ieeetran}
\bibliography{refs}

\end{document}